\documentclass[a4paper,fleqn]{cas-dc}

\usepackage[authoryear]{natbib}

\usepackage[mathscr]{eucal}
\usepackage{amsmath,mathtools}
\usepackage{amssymb}
\usepackage{lipsum}
\usepackage{algorithmicx}
\usepackage[ruled]{algorithm}
\usepackage[noend]{algpseudocode}
\usepackage{subfig}

\usepackage[switch]{lineno} 
\modulolinenumbers[2]

\newcommand*{\img}[1]{%
    \raisebox{0\baselineskip}{%
        \includegraphics[
        height=0.6\baselineskip,
        width=0.6\baselineskip,
        keepaspectratio,
        ]{#1}%
    }%
}

\def\tsc#1{\csdef{#1}{\textsc{\lowercase{#1}}\xspace}}
\tsc{WGM}
\tsc{QE}
\tsc{EP}
\tsc{PMS}
\tsc{BEC}
\tsc{DE}
\begin{document}
\let\WriteBookmarks\relax
\def\floatpagepagefraction{1}
\def\textpagefraction{.001}
\shorttitle{Explaining dimensionality reduction results using Shapley values}
\shortauthors{Marcílio-Jr and Eler.}

\title [mode = title]{Explaining dimensionality reduction results using Shapley values}                      
%\tnotemark[1,2]

%\tnotetext[1]{This document is the results of the research
%   project funded by the National Science Foundation.}

%\tnotetext[2]{The second title footnote which is a longer text matter
%   to fill through the whole text width and overflow into
%   another line in the footnotes area of the first page.}

\author[1]{Wilson E. Marcílio-Jr}
\author[1]{Danilo M. Eler}
%\cormark[1]
%\ead{cvr_1@tug.org.in}
%\ead[url]{www.cvr.cc, cvr@sayahna.org}

%\credit{Conceptualization of this study, Methodology, Software}

\address[1]{Faculty of Sciences and Technology, São Paulo State University (UNESP), Presidente Prudente, SP 19060-900, Brazil}

%\author[2,4]{Han Theh Thanh}[style=chinese]

%\author[2,3]{CV Rajagopal}[%
 %  role=Co-ordinator,
%   suffix=Jr,
%\ead{cvr3@sayahna.org}
%\ead[URL]{www.sayahna.org}

%\credit{Data curation, Writing - Original draft preparation}

%\address[2]{Sayahna Foundation, Jagathy, Trivandrum 695014, India}

%\author%
%[1,3]
%{Rishi T.}
%\cormark[2]
%\fnmark[1,3]
\cortext[cor1]{E-mails: wilson.marcilio@unesp.br, danilo.eler@unesp.br} 
%\ead[URL]{www.stmdocs.in}

%\address[3]{STM Document Engineering Pvt Ltd., Mepukada,
%    Malayinkil, Trivandrum 695571, India}

%\cortext[cor2]{Principal corresponding author}
%\fntext[fn1]{This is the first author footnote. but is common to third
%  author as well.}
%\fntext[fn2]{Another author footnote, this is a very long footnote and
%  it should be a really long footnote. But this footnote is not yet
%  sufficiently long enough to make two lines of footnote text.}

%\nonumnote{This note has no numbers. In this work we demonstrate $a_b$
%  the formation Y\_1 of a new type of polariton on the interface
%  between a cuprous oxide slab and a polystyrene micro-sphere placed
%  on the slab.
%  }

\begin{abstract}
Dimensionality reduction (DR) techniques have been consistently supporting high-dimensional data analysis in various applications. Besides the patterns uncovered by these techniques, the interpretation of DR results based on each feature’s contribution to the low-dimensional representation supports new finds through exploratory analysis. Current literature approaches designed to interpret DR techniques do not explain the features’ contributions well since they focus only on the low-dimensional representation or do not consider the relationship among features. This paper presents ClusterShapley to address these problems, using Shapley values to generate explanations of dimensionality reduction techniques and interpret these algorithms using a cluster-oriented analysis. ClusterShapley explains the formation of clusters and the meaning of their relationship, which is useful for exploratory data analysis in various domains. We propose novel visualization techniques to guide the interpretation of features’ contributions on clustering formation and validate our methodology through case studies of publicly available datasets. The results demonstrate our approach’s interpretability and analysis power to generate insights about pathologies and patients in different conditions using DR results. 
\end{abstract}

% \includegraphics[width=\linewidth]{figs/visualization-design/tool.pdf}

% \begin{highlights}
% \item Clusters of dimensionality reduction results can be interpreted using Shapley values
% \item Summary visualizations convey
% \item Novel visualization methods are presented to analyze classifier confusion
% \end{highlights}


\maketitle

% \linenumbers

\section{Introduction}
\label{sec:introduction}

Dimensionality reduction (DR) techniques help analyze high-dimensional datasets by mapping data from high dimensions ($\mathbb{R}^m$) to data in low dimensions ($\mathbb{R}^d$). These techniques try to preserve, as much as possible, the relationship among data samples present in the original space ($\mathbb{R}^m$). Thus, researchers employ scatter plot-based representations in exploratory analysis to look for patterns and other relevant information in data. There are several examples of studies DR techniques for exploratory data analysis, such as understanding the learned features by CNNs during different epochs~\citep{Pezzotti2018} or investigating gene expression patterns to discover new cell types~\citep{Unen2018} and many others. Although DR techniques offer an excellent opportunity for high-dimensional data analysis, analysts must interpret the decisions made by these algorithms to understand if the DR results encode the information in the high-dimensional space. For example, understanding the DR result helps machine learning practitioners assess the quality of feature spaces regarding class separation~\citep{Marcilio2020}.

For the interpretation of dimensionality reduction techniques, one possible solution consists of analyzing the features' contributions to the DR result, assessing how much each feature contributed to forming clusters or other visible structures in the projected space (e.g., $\mathbb{R}^2$). For example, in gene expression analysis, bioinformaticians want to know which genes influence each cluster to annotate cell types or discover new ones~\citep{Lahnemann2020}. Finding the contribution of these features to the dimensionality reduction result is mainly related to non-linear DR techniques~\citep{Maaten_2008, McInnes2018}, in which there is no current way to inverse calculations and keep track of feature contributions during algorithm execution~\citep{Fujiwara2019}. Nevertheless, non-linear DR techniques are the most suitable for dealing with most datasets due to their ability to uncover complex structures.

Existing techniques for DR interpretation present a few problems. For example, works that focus on feature values~\citep{Coimbra2016,Pagliosa2016,Marcilio2020} do not account for the dimensionality reduction process and only focus on the reduced low-dimensional space. Other more elaborated works~\citep{Turkay2012,Joia2015} obtain feature importance through principal components (PC). Using the PCs returns biased outputs for classes with high variation~\citep{Joia2015}, and their inability to focus on local information~\citep{Fujiwara2019} impairs cluster-oriented analysis. A robust approach, called ccPCA~\citep{Fujiwara2019}, uses contrastive PCA~\citep{Abid2018} to understand DR results based on each cluster's specific information. Using contrastive analysis, ccPCA emphasizes what is different from each cluster. For example, for a dataset of machine learning papers talking about classification, it would return the information that differentiates them, such as the classification method. These techniques cannot explain how much each feature contributed to the DR result. The importance measure assigned to the features does not capture their contribution to clusters and other structures. More importantly, these feature importance measures do not interact with each other to construct an explanation measure. Instead, the importance of each feature is independent of the other.

In this work, we push to the state-of-the-art problem of interpreting dimensionality reduction results. More specifically, we propose a novel methodology to explain the feature contributions in cluster formation in dimensionality reduction results. Using Shapley values~\citep{Shapley1953} to derive explanations, we interpret DR results using the features' contributions in an addictive way to show how much each feature contributes to the resulting projection in the visual space ($\mathbb{R}^2$). Besides explaining DR results and supporting data sample analysis using the similarity among Shapley values, our methodology allows the extrapolation from feature contribution to feature importance concerning cluster formation. Our method, called ClusterShapley, consists of a novel application of Shapley values, and it helps analysts to understand the decisions of DR techniques after projection. Finally, we also propose summary visualizations to depict Shapley values, and Kernel Density Estimation~\citep{Rosenblatt1956,Parzen1962} to aggregate highly correlated feature contributions.

 In the case studies, we show how the feature contributions can explain interesting patterns and reveal insights about medical and social datasets. Then, we discuss the implications of our work by delineating possible applications using ClusterShapley. Finally, we emphasize this is the first research study using Shapley values to explain dimensionality reduction results in a cluster-oriented analysis. Summarily, the contributions of this paper are:

\begin{itemize}
\item A methodology for applying Shapley values to explain dimensionality reduction results upon a cluster-oriented analysis;
\item Summary visualizations to encode feature contributions based on Shapley values;
\item Categorization of important features in datasets about pathologies and patients in different conditions.
\end{itemize}

This paper is organized as follows: we delineate the related works in Section~\ref{sec:related-works}, a brief background on Shapley values is provided in Section~\ref{sec:background}, we explain our methodology and the visualization design in Section~\ref{sec:methodology}, validation of our approach is performed through case studies in Section~\ref{sec:case-studies}, a discussion is provided in Section~\ref{sec:discussions}, we conclude our work in Section~\ref{sec:conclusion}.

\section{Related Works}
\label{sec:related-works}

To improve interpretation capabilities of dimensionality reduction (DR) techniques, researchers provide additional information to these methods' results. Usually, a few works include visual information such as bar charts and color encoding to interpret three-dimensional projections~\citep{Coimbra2016} or encode attribute variation using Delaunay triangulation to assess neighborhood relations in two-dimensional projections~\citep{Silva2015}. Probing Projections~\citep{Stahnke2016}, for example, depicts error information by dis-playing a halo around each dot in a DR layout besides providing interaction mechanisms to understand distortions in the projection process. The majority of the works use traditional statistical charts to visualize attribute variability~\citep{Pagliosa2016}, neighborhood and class errors~\citep{MarcilioJr2017}, or quality metrics~\citep{Kwon2018}.

More related to our work are the techniques that try to find important features given clusters of data samples. For example, the Linear Discriminative Coordinates~\citep{Wang2017} use LDA~\citep{Izenman2008} to produce cohesive clusters by discarding the least important features.~\citet{Joia2015} use PCA to find the most important features by a simple matrix decomposition and visualize feature names as word clouds within each cluster region. Although useful and fast, classes with high variation influence the result. Another work, proposed by~\citet{Turkay2012}, also uses PCA's principal components to obtain the representative features of a multidimensional scaling \citep{Kruskal1978} projection. Recently,~\citet{Fujiwara2019} proposed a cluster contrastive PCA (ccPCA) technique that finds the most important features for a given cluster in contrast with the other clusters in a DR result. Fujiwara et al.'s approach is different from Joia et al.'s and Turkay et al.'s works. It provides a way to understand which features positively contribute to the differentiation of clusters.

These works cannot explain how much a feature contributes to the dimensionality reduction result. That is, the feature importance does not show its contribution to the dimensionality reduction result.

Our work adds to the state-of-the-art interpreting dimensionality reduction results by showing how each feature contributes to the results. It explains the position of the data samples on the DR result through the combination of each feature's contribution.

\section{Review of Shapley values}
\label{sec:background}

One crucial consideration when explaining machine learning models is how to explain predictions without taking the model itself into account. As discussed by~\citet{Strumbelj2014}, the critical component of a model-agnostic explanation consists of each feature's contributions to the prediction. In other words, the influence of each feature of the dataset explains a prediction. Let $f$ be a machine learning model, and $x$ be an instance from a dataset $X$. The situational importance~\citep{Achen1982} of the feature (Equation~\ref{eq:situational-importance}) computes how a particular value influences a prediction. It consists of the difference between the feature contribution when its value is $x_i$ and its expected contribution~\citep{Strumbelj2014}, where $\beta_i$ denotes the $i$th feature's global importance.

\begin{equation}
    \label{eq:situational-importance}
\varphi_i(x) = \beta_ix_i - \beta_iE[X_i].
\end{equation}

\noindent For feature $i$, the situational importance signal indicates positive, negative, or no contribution. Although such contributions are easy to compute for addictive models (such as linear regression models), it is a difficult task for complex models due to the interaction among features. The conditional expectation of a model’s prediction~\citep{Strumbelj2014}, defined in Equation~\ref{eq:situational-importance-subset}, takes all subsets of features ($Q$) into account.

\begin{equation}
    \label{eq:situational-importance-subset}
f_Q(x) = \mathop{E}[f|X_i = x_i, \forall_i \in Q].
\end{equation}

\noindent Shapley values – a concept from coalitional game theory~\citep{Shapley1953} – can be used~\citep{Lundberg2017} to measure each feature's contribution to the prediction of a model using Equation~\ref{eq:situational-importance-subset}. Shapley values are computed by averaging each feature permutation on the conditional expectation of a model prediction, or in other words, measuring the change in the prediction after adding a feature into Equation~\ref{eq:situational-importance-subset}. Thus, in the equation Equation~\ref{eq:shapley-value}~\citep{Lundberg2020}, $Q$ represents all feature permutations, $y_i^q$ is the set of all features that come before feature $i$ in the permutation $q$, $y_i^q \cup i$ is the union of the set of all features that come before feature $i$ in the permutation $q$ and the feature $i$ itself, and $|Y|$ corresponds to the number of input features for the model.

\begin{equation}
\label{eq:shapley-value}
\phi_i(f, x) = \sum\limits_{q \in Q} \dfrac{1}{|Y|!}(f_{(y_i^q \cup i)}(x) - f_{(y_i^q)}(x))
\end{equation}

\noindent For a dataset with $n$ features, $2^n$ model evaluations would be necessary to compute Shapley values, making it prohibitive for moderate numbers of $n$. So, in this work, we use KernelSHAP~\citep{Lundberg2017} to approximate Shapley values using a linear regression model.

\subsection{Shapley values exemplification}

To illustrate how Shapley values help understand the contributions to a model’s prediction, we could think about a synthetic scenario~\citep{Molnar2019}. Suppose we trained a machine learning model to predict house prices, and for a particular house, it predicts $300$ thousand dollars. For such a prediction, our model used the following features: \texttt{pets allowed}, \texttt{one-bedroom}, \texttt{size} of $100$ $m^2$, and two \texttt{bathrooms}.

The average prediction of our synthetic scenario was $290$ thousand dollars. For our particular example, Shapley values tell us how much each feature contributed to the prediction compared to the average. Shapley values explain the difference between the contributions for the prediction ($300$ thousand dollars)~\citep{Molnar2019} and the average prediction (290 thousand dollars), which is $10$ thousand dollars. In the end, one possible solution for this problem could be: \texttt{pets allowed} contributed $40$ thousand dollars, \texttt{one-bedroom} contributed -$60$ thousand dollars, \texttt{size} of $100$ $m^2$ contributed $10$ thousand dollars, and two bathrooms with $20$ thousand dollars. Notice that these values sum to $10$ thousand dollars.

Now, following the idea discussed to formulate Equation~\ref{eq:shapley-value}, the Shapley value for a particular feature consists of the average contribution of such a feature according to all possible feature permutations. To estimate the Shapley value for the \texttt{pets allowed} feature, one has to compute the prediction with: \texttt{pets allowed}; \texttt{pets allowed} and \texttt{size} of $100$ $m^2$; \texttt{pets allowed}, \texttt{size} of $100$ $m^2$, and \texttt{two bathrooms}, and so on. One particular important thing is that the feature \texttt{pets allowed} also has to be removed for each permutation. Finally, by removing a feature for the prediction, we mean to assign a random value (from another data sample) to the ``removed'' feature. Also, one can get a better estimation when repeating the sampling process and contributions averaged.

Due to its exponential nature, the application of Shapleyvalues is infeasible for high-dimensional datasets. Thus, as discussed previously, we use a technique called KernelSHAP that estimates Shapley values through a linear regression upon sampled permutations. 

\section{Using Shapley values to explain clusters of dimensionality reduction results}
\label{sec:methodology}

In this section, we explain how to define clusters for computing the explanations using Shapley values. Then, we delineate the visualization system used to help analysts with hypothesis generation.

\begin{figure}[h!]
\centering
\includegraphics[width=\linewidth]{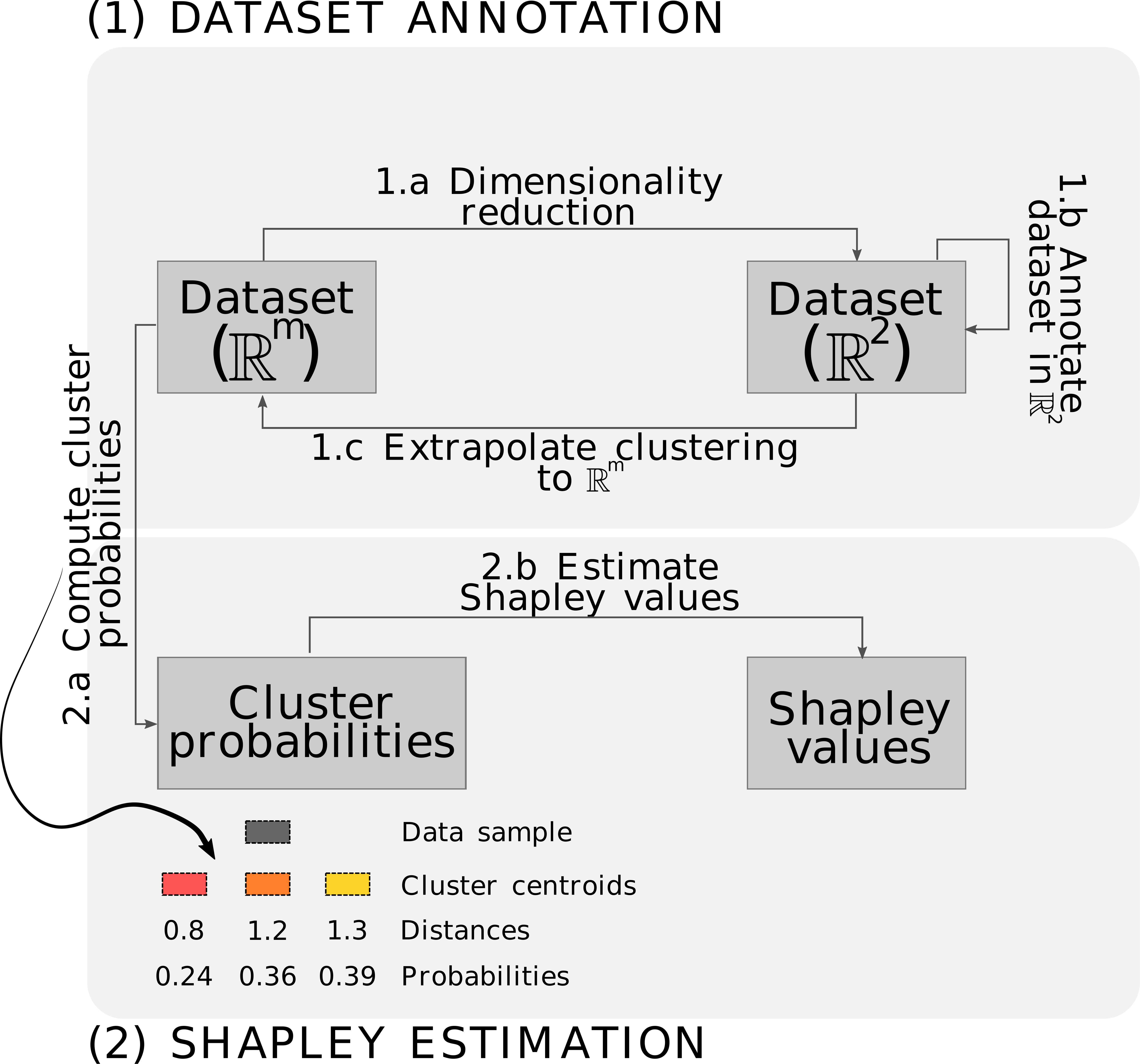}
\caption{ClusterShapley framework. In the ``dataset annotation'' component (1), a dimensionality reduction (1.a) process helps users to analyze and to annotate clusters perceived in the visual space ($\mathbb{R}^2$); then, the clusters defined in $\mathbb{R}^2$  annotate the high-dimensional dataset (1.c). In the ``Shapley estimation'' component (2), the annotated high-dimensional dataset is used to generate clusters probabilities for each data sample (2.a) employed for Shapley values estimation (2.b).}\label{fig:framework}
\end{figure}

Figure~\ref{fig:framework} shows the two main components to derive explanations after dimensionality reduction. The ``Dataset annotation'' (1.) component defines the clusters explained in the second component, ``Shapley estimation'' (2.). Notice that, instead of showing the Shapley values as a measure of importance to the dimensionality reduction result, we provide novel visual metaphors to facilitate analysis and exploratory analysis. 

\subsection{Dataset annotation}
\label{sec:dataset-annotation}

To explain dimensionality reduction (DR) techniques, we rely on interpreting the clusters formed on the projected space ($\mathbb{R}^2$). DR techniques aim to reduce the dimensionality of a high-dimensional dataset ($\mathbb{R}^m$) to a low-dimensional dataset ($\mathbb{R}^d$) while preserving (as much as possible) the structures present in the data. We usually use $d=2$ to visualize the result of a DR technique, in which the visual proximity among data points encodes similarity. For instance, clusters are rapidly perceived in the visual space ($\mathbb{R}^2$) since humans quickly notice groups of visual markers~\citep{Bertin_1983}.

\begin{figure}[!htp]
\centering
\includegraphics[width=0.5\linewidth]{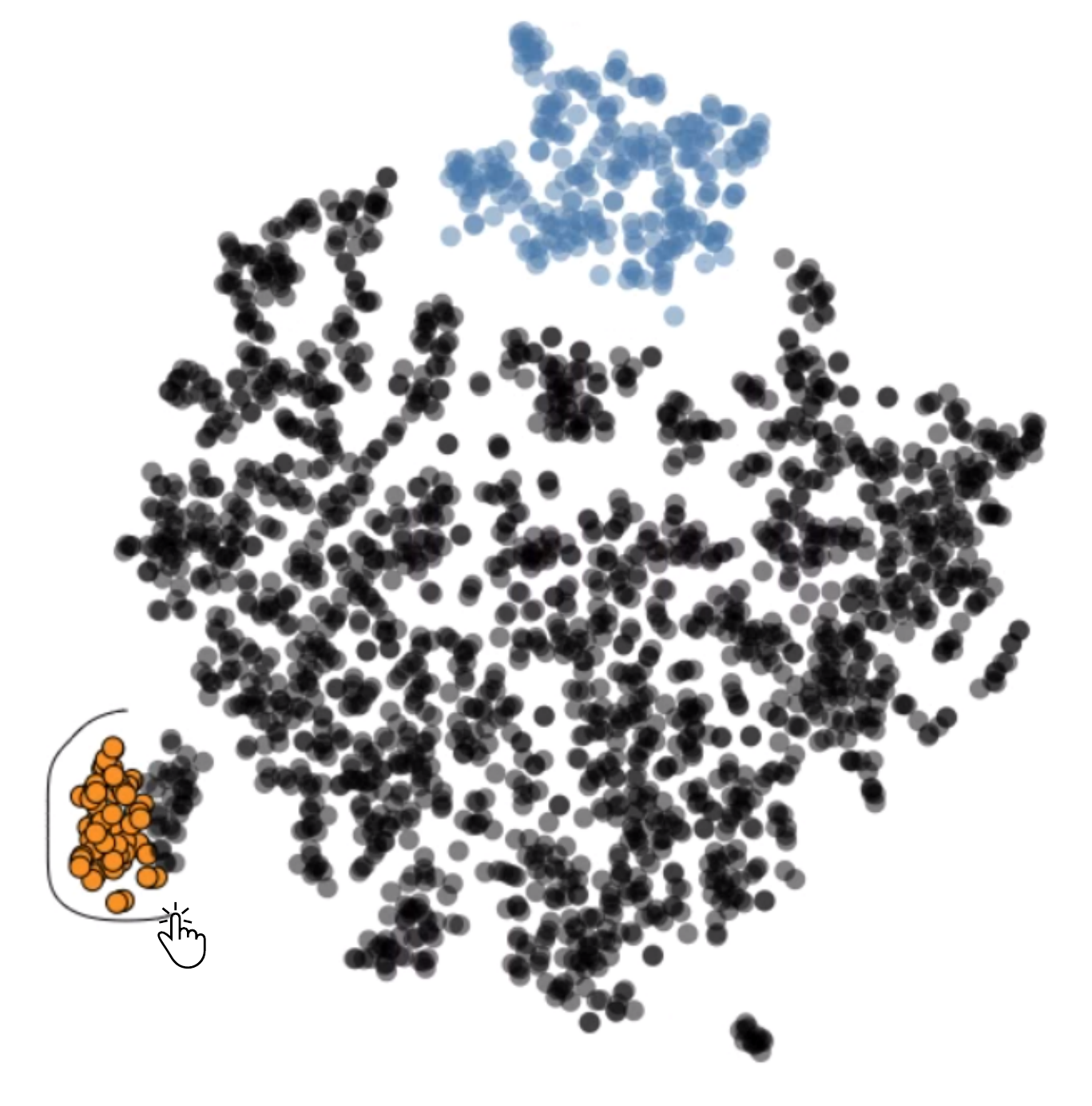}
\caption{Cluster definition through manual selection. Users receive a projected dataset with no clustering imposed, and then, users define clusters by lassoing the groups perceived in the visual space. Notice that black circles indicate data samples not belonging to any cluster, while circles with different data hues indicate already assigned data samples.}\label{fig:defining-clusters}
\end{figure}

To understand the DR technique's decisions to produce the projection, we use the clusters on $\mathbb{R}^2$ to annotate the dataset in the high-dimensional space. In this case, users can freely define clusters with mouse interaction, as shown in Figure~\ref{fig:defining-clusters} – where black color encodes data samples not assigned to any cluster. Notice that by using such an approach, users might select data samples of different classes projected on the same cluster when using labeled datasets. Such an idea is reasonable and consistent with our proposal since we want to understand and explain the visual space clusters. Section~\ref{sec:indian-liver} presents a study case where we manually annotate clusters on data with mixed classes.

Other possible ways to define clusters are to precompute a clustering algorithm~\citep{KaufmanRousseuw2005} or to use labeled datasets. Using one of these three strategies, we aim to define as many clusters as the number of clusters perceived in the visual space. For example, for the projection of Figure~\ref{fig:defining-clusters}, one may define seven clusters (as shown for a study case Section~\ref{sec:communities-crime}).

Algorithm~\ref{alg:annotate-clusters} shows the steps performed to annotate clusters. We receive a high-dimensional dataset $X \in \mathbb{R}^m$, the data annotation \texttt{method}, and the arguments (\texttt{args}) for the techniques. There is nothing to do for labeled datasets (Line 2), and we return the annodated high-dimensional data (Line 3). Suppose the annotation method is \texttt{clustering} or \texttt{manual} annotation. In that case, we have to reduce the dimensionality of the dataset to $\mathbb{R}^2$ (Line 4) so that analysts can investigate the clusters in the visual space. Users either run a clustering algorithm (Line 6) or use mouse interaction to manually define the clusters (Line 8). The labels produced by one of these methods annotate the high-dimensional dataset in Lines 9 and 10.

\begin{algorithm}[]
\small
\caption{Annotating dataset.}
\label{alg:annotate-clusters}
\begin{algorithmic}[1]
\Procedure{dataset\_annotation}{$X\in\mathbb{R}^m$, method, args}
    \If{method = 'labeled\_dataset'}
        \State \Return X
    \EndIf
    \State $X'\leftarrow$ DimensionalityReduction($X$, args)
    \If{method = 'clustering'}
        \State labels $\leftarrow$ Clustering($X'$, args)
    \EndIf
    \If{method = 'manual'}
        \State labels $\leftarrow$ Interaction($X'$, args)
    \EndIf
	\For{$i \in |X|$} 
	    \State $X_i.$label $\leftarrow$ labels$_i$
	\EndFor
	\State \Return $X$
	
\EndProcedure
\end{algorithmic}
\end{algorithm}

\subsection{Shapley estimation}
\label{sec:shapley-estimation}

After the cluster definition, we can generate Shapley values for each data sample. Thus, we need to define a model $f$ that returns the prediction probabilities for a data sample $x$ based on the cluster definition -- discussed in the previous section.

To return the prediction probabilities for a data sample $x$, we measure the distance from $x$ to each cluster centroid. Figure~\ref{fig:framework} (2.a -- bottom) illustrates such a process for three cluster centroids (\img{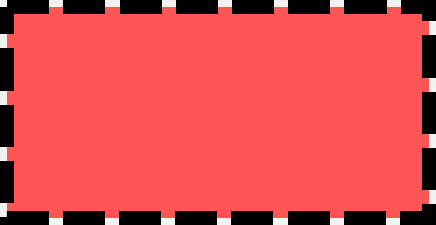}, \img{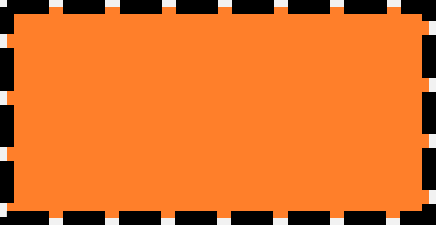}, \img{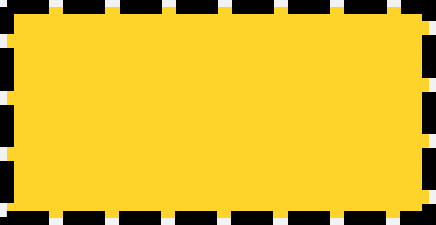}) and, consequently, a three-dimensional distance vector. To convert the distances into probabilities, we apply an L1 normalization. The Shapley estimator (in our case, KernelSHAP~\citep{Lundberg2017}) uses these probabilities (for each data sample) to generate explanations discussed in Section~\ref{sec:background}. Notice that while estimating the Shapley values using KernelSHAP accounts for most of the dataset, we only compute the estimation for 20\% of the data. The result of this procedure will be a matrix of dimensions $n\times M$ for each cluster, where $n$ corresponds to 20\% of the dataset size and $M$ represents the dimensionality of the dataset – each cell $i,j$ of the matrix will contain the Shapley value of the datapoint $i$ for the feature $j$.

Algorithm~\ref{alg:shapley-estimation} further illustrates the Shapley values estimation process. First, we call the estimation procedure with the annotated high-dimensional dataset ($X \in \mathbb{R}^2$). Then, we split the dataset into training and test sets (Line 8) to compute Shapley values for the test set (Line 1) using the training set to fit the algorithms (Line 9). To create an instance for Shapley estimation, we have to provide a function that will return the prediction probabilities. Such a function (cluster\_probability) uses the clusters' centroids (Line 2) to compute the distance from a data sample $x$ to these centroids (Lines 4 and 5) and then returns the prediction probabilities in Line 6.

\begin{algorithm}[]
\small
\caption{Shapley estimation.}
\label{alg:shapley-estimation}
\begin{algorithmic}[1]

\Procedure{cluster\_probability}{$x\in\mathbb{R}^m$, $X\in\mathbb{R}^m$}
    \State $C\leftarrow$ get\_centroids($X$)
    \State $D\leftarrow \emptyset$
    \For{$i\in |C|$}
        \State $D\leftarrow D \cup\ (\parallel x-C_i\parallel)$
    \EndFor
	\State \Return $D/\parallel D\parallel_1$
	
\EndProcedure

\Procedure{shapley\_estimation}{$X\in\mathbb{R}^m$}
    \State $X_{train}, X_{test}\leftarrow$ split($X$, $0.2$)
    \State SE $\leftarrow$ KernelSHAP(cluster\_probability, $X_{train}$)
    \State SV $\leftarrow$ SE($X_{test}$)
	\State \Return $SV$
	
\EndProcedure
\end{algorithmic}
\end{algorithm}

An essential consideration of the L1 normalization is that lower probabilities will indicate better cohesion with clusters. As we will see in the following section, negative Shap-ley values indicate that a feature contributed to the cluster cohesion, while positive Shapley values influence the non-cohesion of clusters. Such a characteristic fits well for the visual space, where visual proximity encodes similarity.

Finally, we emphasize that any technique can perform the dimensionality reduction process, such as t-SNE \citep{Maaten_2008}, LSP \citep{Paulovich_2008}, or UMAP \citep{McInnes2018}. Different clusters may be perceived in the visual space when using different dimensionality reduction techniques. Thus, labeling these clusters (see Section~\ref{sec:dataset-annotation}) would be the first step to apply our methodology.

The estimated Shapley values correspond to each feature's contribution to the dimensionality reduction result. Thus, a feature with a high absolute Shapley value contributes a lot to the projected dataset cluster formation. In this case, each data point used for Shapley values estimation contains the correspondent Shapley value. Negative Shapley values mean that a feature contributes to the cluster formation, and positive Shapley values mean that a feature does not contribute to cluster formation. The following section provides novel visualization approaches to interpret dimensionality reduction results using the estimated Shapley values.

\section{Visualization Design}
\label{sec:visualization-design}

\begin{figure}[h!]
\centering
\includegraphics[width=\linewidth]{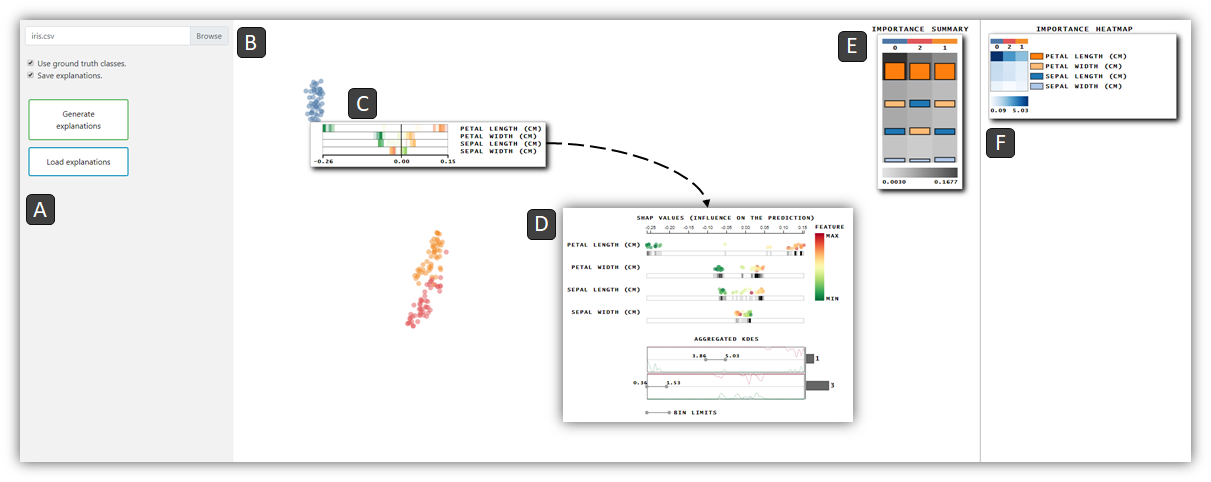}
\caption{Overview of the prototype system. Datasets can be specified in (A), where users can also load previously generated explanations. The scatter plot representation is drawn in (B), where we use color to encode classes. Summary representations of Shapley values, used to indicate importance, are given by hovering particular class instances (C). The detailed analysis of Shapley values and feature importance helps analysts understand the correlation between feature value and Shapley value (D). In (E), we show the \textbf{Importance Summary}, which summarizes the mean of the absolute Shapley values for each pair (class, feature). A heatmap with the sum of the absolute Shapley values shows the importance overview (F).}\label{fig:overview-iris}
\end{figure}

Figure~\ref{fig:overview-iris} shows a prototype system using ClusterShapley for explaining a dimensionality reduction result for the \textit{Iris}~\citep{Dua2019} dataset. The system has three components. In the first component (A), users can provide datasets to generate explanations and load stored explanations. The scatter plot representation of DR results is drawn in the second component (B). Finally, the third component corresponds to the visualizations provided to support DR results using Shapley values. The tool-tip (C) next to the blue cluster shows a visual summarization of the Shapley values for the four most important features. Users can also click on a circle to show a detailed analysis of the Shapley values, as shown in (D). The detailed analysis is based on a dot plot and aggregated Kernel Density Estimation~\citep{Rosenblatt1956,Parzen1962} of the absolute sum of Shapley values. The \textbf{Importance Summary} is shown in (E), where users assess the mean values for each class's four most important features and the contribution of these features for characterizing the clusters depicted by bar-plots. Finally, we also provide a heatmap of the sum of Shapley values in (F). The following sections present details of each component.

\subsection{Scatter plot component}

To provide a rapid assessment of Shapley values, we use a tool-tip with summarized information about the interactions among feature values and Shapley values when users hover circles of a particular class. The process of summarizing the Shapley values and features values works as follows. $\mathscr{H}^i_e$ and $\mathscr{H}^i_o$ are the histograms created from the Shapley values for the feature $i$, where $\mathscr{H}^i_e$ stores information for values equal or greater than the mean ($\sigma_i$) and $\mathscr{H}^i_o$ stores information lower than the mean of feature values for feature $i$. Knowing that Shapley$_{min}$ and Shapley$_{max}$ consist of the lowest and the greatest Shapley values for the visualized features, we divide the histograms $\mathscr{H}^i_e$ and $\mathscr{H}^i_o$ in the same number of bins using Shapley$_{min}$ and Shapley$_{max}$ as bin limits.

\begin{figure}[h!]
\centering
\includegraphics[width=\linewidth]{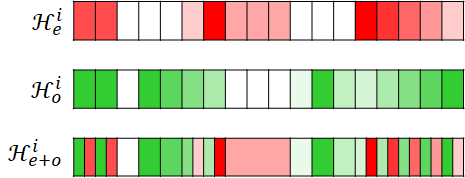}
\caption{Summarization approach to encoding two histograms for a given feature $i$, $\mathscr{H}^i_e$ for feature values equal or greater than the mean ($\sigma_i$) and $\mathscr{H}^i_o$ for features values lower than the mean. The two histograms are encoded together by alternating their bins.}\label{fig:summarization-scheme}
\end{figure}

Color saturation encodes each histogram's density -- green for $\mathscr{H}^i_e$ and red for $\mathscr{H}^i_o$ -- and the two histograms have their bins sequentially drawn one next to another. That is, while $\mathscr{H}^i_e$ takes the even positions, $\mathscr{H}^i_o$ takes the odd ones. Figure~\ref{fig:summarization-scheme} illustrates this process. If only one histogram has a density greater than zero for a given bin, it will use the even and odd positions.

\begin{figure}[h!]
\centering
\includegraphics[width=\linewidth]{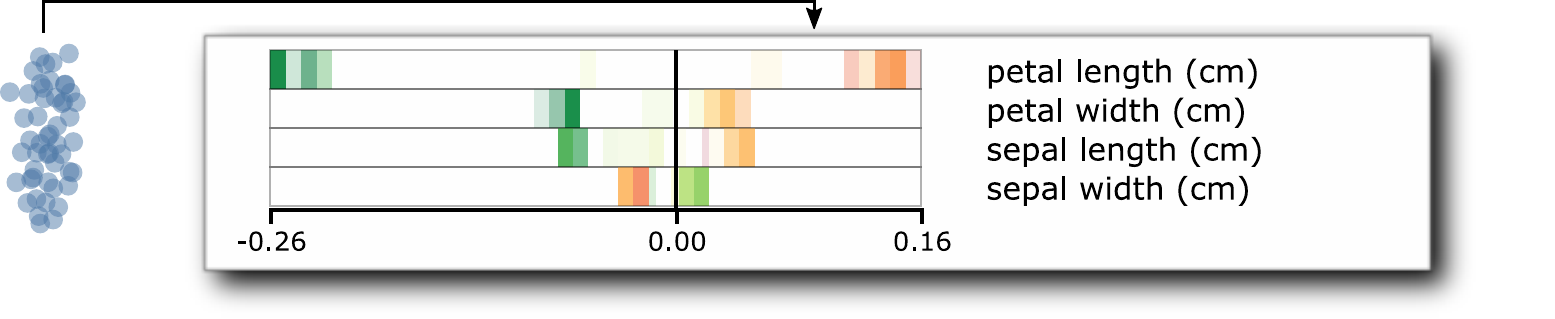}
\caption{A tool-tip for summarizing information of the blue cluster of Figure~\ref{fig:overview-iris}.}\label{fig:summarization-example}
\end{figure}

On real data, the summarized information looks like in Figure~\ref{fig:summarization-example}, where reddish colors encode feature values greater than the mean and greenish colors encode feature values lower than the mean. The contribution of the feature value is encoded using position, which corresponds to the Shapley values -- the contribution is proportional to the distance from $0$, indicated by a vertical line segment. The ordering in the representation indicates overall feature importance, i.e., \texttt{petal length (cm)} is the most important, \texttt{petal width (cm)} is the second most important, and so on.

\subsection{Coordination between scatter plot and Shapley values}

\begin{figure}[h!]
\centering
\includegraphics[width=\linewidth]{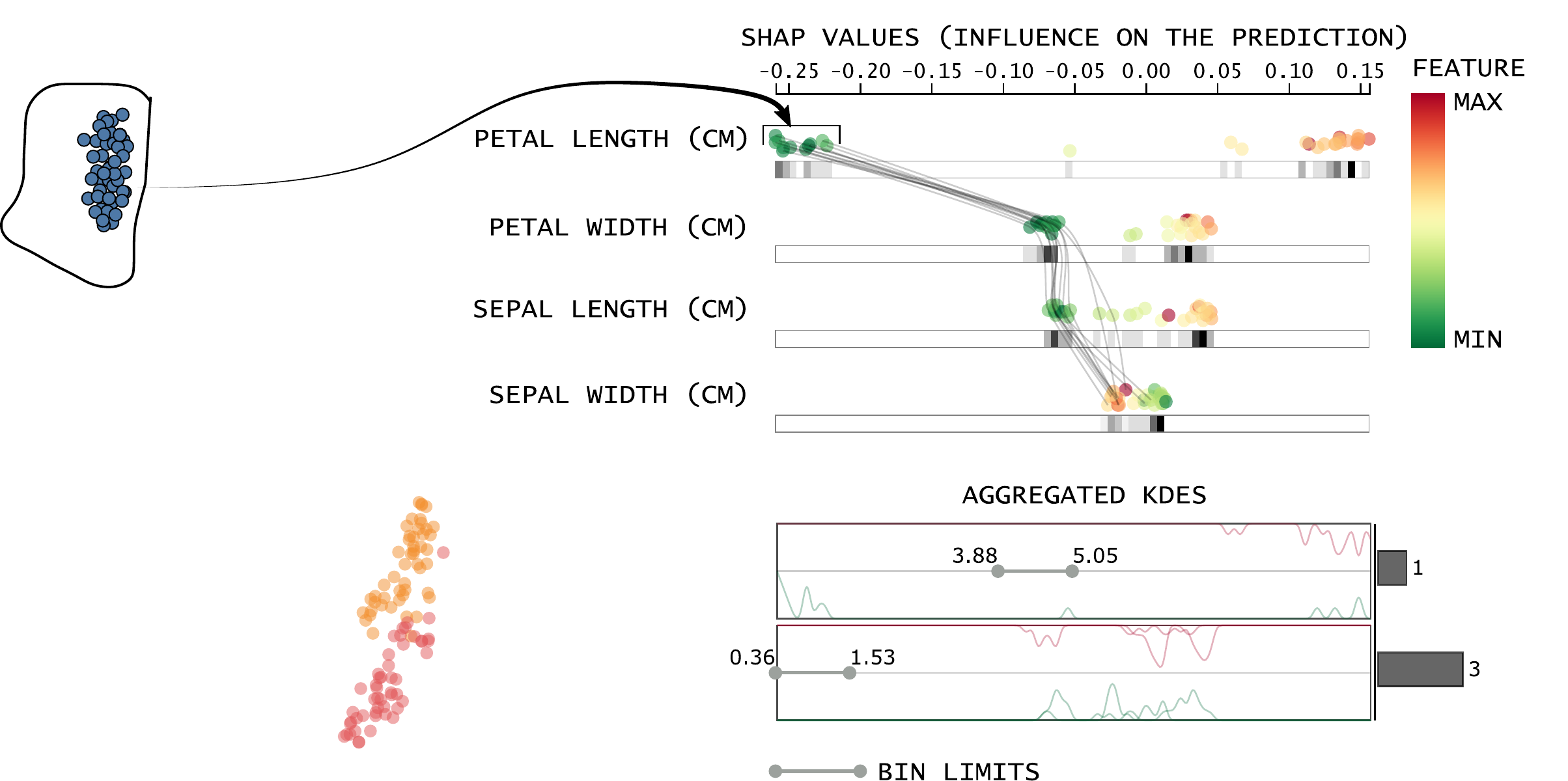}
\caption{Coordination between the scatter plot and the dot plot encoding the Shapley values. Each line segment represents a point in the scatter plot representation.}\label{fig:interaction-coordination}
\end{figure}

Analyzing the correlation among feature values and Shapley values is essential for our scenario. We provide a detailed analysis when users click on a circle representing a particular class. Such a detailed analysis is supported by dot plots of the four most important features, as shown in Figure~\ref{fig:interaction-coordination}. For instance, the four most important features ordered according to their importance are \texttt{petal length (cm)}, \texttt{petal width (cm)}, \texttt{sepal length (cm)}, and \texttt{sepal width (cm)}. Notice that as the absolute Shapley values (encoded as horizontal position) assume values far from zero, the more influence the feature will have in characterizing the cluster. So, the density bars below each dot plot help assess how much the feature values contribute to the cluster characterization. 

In the dot plot representation, we visualize the subset of data samples used for computing Shapley values. Each circle encodes a feature value of a data point. While color encodes the feature value, the points' position encodes the Shapley values of the inspected cluster. Lastly, suppose the dataset has more than four features. In that case, they can also be inspected based on the absolute sum of their Shapley values using a histogram, as shown in Figure~\ref{fig:interaction-coordination} (\textbf{Aggregated KDEs}). We draw only the bins with elements (specified by the bin limits), while the bars encode how many features are in the corresponding bin limit. For instance, in Figure~\ref{fig:interaction-coordination}, there is one feature inside $[3.86, 5.03)$ and three features inside $[0.36, 1.53)$. Two Kernel Density Estimation (KDE) curves encode each bin's aggregation: the red curves encode the feature values equal or greater than the mean, and the green curves encode the feature values lower than the mean. Here, we have \texttt{petal length (cm)}, inside the bin with limits $[3.86, 5.03)$ and \texttt{petal width (cm)}, \texttt{sepal length (cm)}, and \texttt{sepal width (cm)} inside the bin with limits $[0.36, 1.53)$.

%We use dot plots for the four most important features together with a density information encoded as color intensity. Note that as the absolute Shapley value gets far from zero, more influence the feature will have in characterizing the cluster, so that, the density helps users to assess how much the dots weight for a particular feature. In addition, features ordering is used to convey their importance for the clustering, i.e., features on top are the most important. Lastly, we also convey the remaining features using Kernel Density Estimation (KDE). For this, the features are aggregated based on the absolute sum of their Shapley values -- as depicted by the bin limits --, then, after computing a histogram (we used bin size $= 100$), KDE is used to generate the curves and to provide a summary about features contribution using Shapley values. 

For the detailed inspection, The coordination mechanism between the scatter plot and the Shapley values helps users identify which features contributed to the cluster cohesion and which features contributed to dispersing the clusters. Figure~\ref{fig:interaction-coordination} illustrates the result of these operations, where we highlight in the dot plot representation the data points selected in the scatter plots.

%As a similar approach in the star plot summarization of feature values proposed by Marcilio-Jr and Eler~\cite{MarcilioJr2020}, range selection is allowed to find patterns among feature values when selecting various data instances -- the result is similar as presented in Figure~\ref{fig:interaction-coordination}. 

\subsection{Importance Summary and Importance Heatmap}

In the \textbf{Importance Summary} (see Figure~\ref{fig:overview-iris} (E)), color intensity depicts the mean of absolute Shapley values for the four most important features. Bar height depicts the contribution of each feature, where color encodes the features. By positioning the features bar with different colors, users assess the heterogeneity of the importance of feature importance for different clusters.

The \textbf{Importance Heatmap} (see Figure~\ref{fig:overview-iris} (F)) provides an overview of the features' contribution by showing the Shapley values' absolute sum for each feature. After computing the absolute sum for each pair of feature and cluster, we use Fujiwara et al.'s~\citep{Fujiwara2019} approach to order the cells to facilitate cluster identification. That is, we apply a hierarchical clustering~\citep{Mullner2011} on rows and columns of the heatmap. Then, optimal-leaf-ordering~\citep{BarJoseph2001} orders the clustering leaves to give more understandable results, positioning similar heatmap values appear close to each other.

The main difference between our heatmap to Fujiwara et al.'s~\citep{Fujiwara2019} is how we summarize information for datasets with too many features. We find the most important features for each cluster, showing at most $min(4, m)$ features, where $m$ denotes the dataset's dimensionality.

\subsection{Analysis example}

To make readers familiar with the analysis using ClusterShapley, we inspect the explanations for a dimensionality reduction result on the Iris dataset. Moreover, we use an icon (such as \img{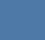}) and the cluster's indices to facilitate reading. 

Figure~\ref{fig:overview-iris} shows a separation of cluster $0$~\img{figs/0.png} from the others. The \textbf{Importance Summary} shows that \texttt{petal length} has much more influence on the cluster. The summary visualization of Shapley values in Figure~\ref{fig:overview-iris} (C) gives a hint of how feature values influence the cluster formation, i.e., there is a clear separation of lower values and higher values for \texttt{petal length}. In Figure~\ref{fig:interaction-coordination}, the coordination between the scatter plot and dot plot shows that due to lower values of \texttt{petal length}, \texttt{petal width}, and \texttt{sepal length}, instances of cluster $0$ \img{figs/0.png} are very different from the others. The negative Shapley values show that their feature values contributed to the cluster cohesion.

\begin{figure}[!htb]
    \centering
    \subfloat[Feature values for cluster orange seems to be moderate (yellowish colors).]{\includegraphics[width=\linewidth]{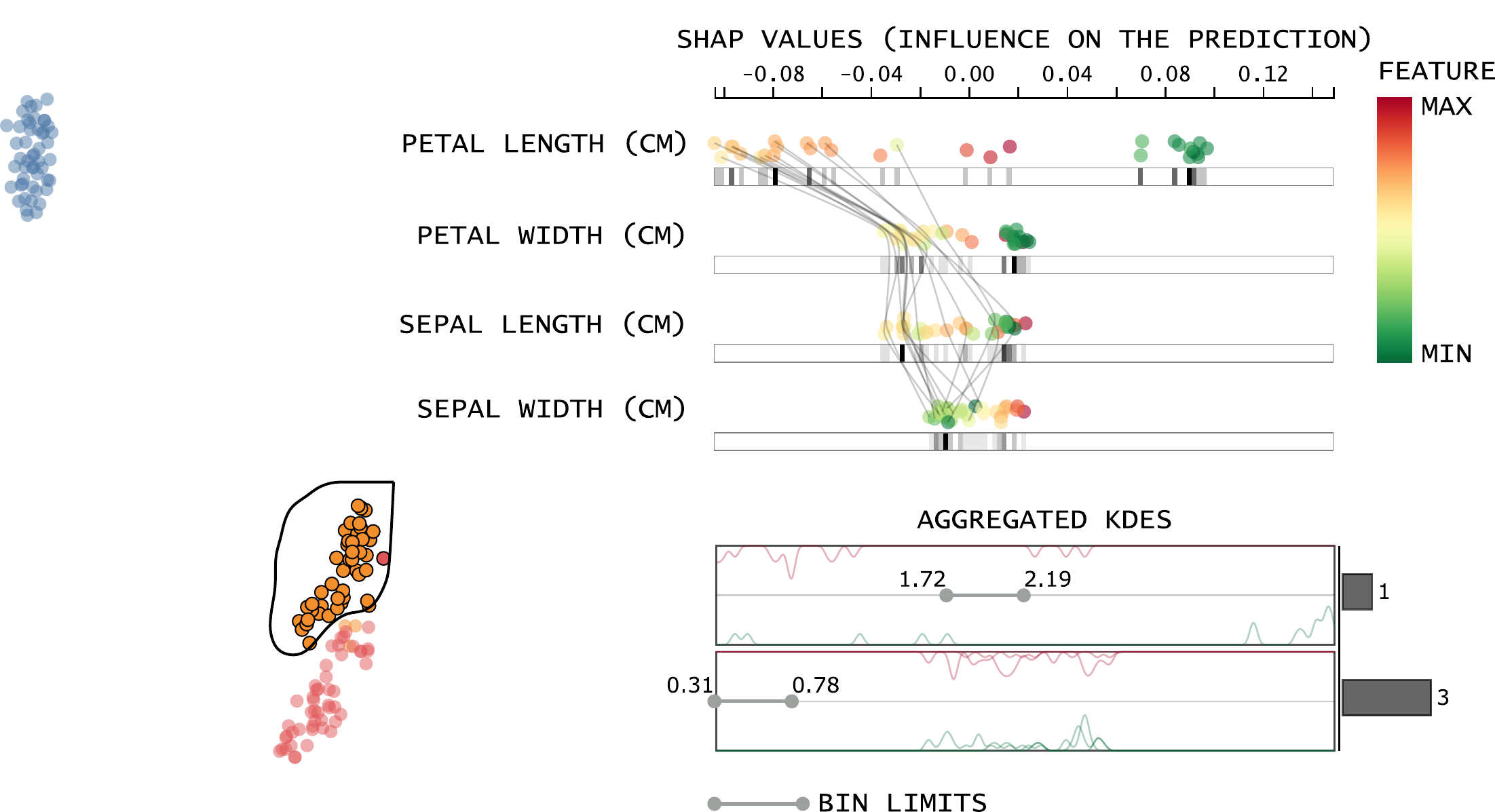}\label{fig:iris_dotplot-class1}}

    \subfloat[Feature values for cluster red seems to be higher if compared with the features of orange cluster.]{\includegraphics[width=\linewidth]{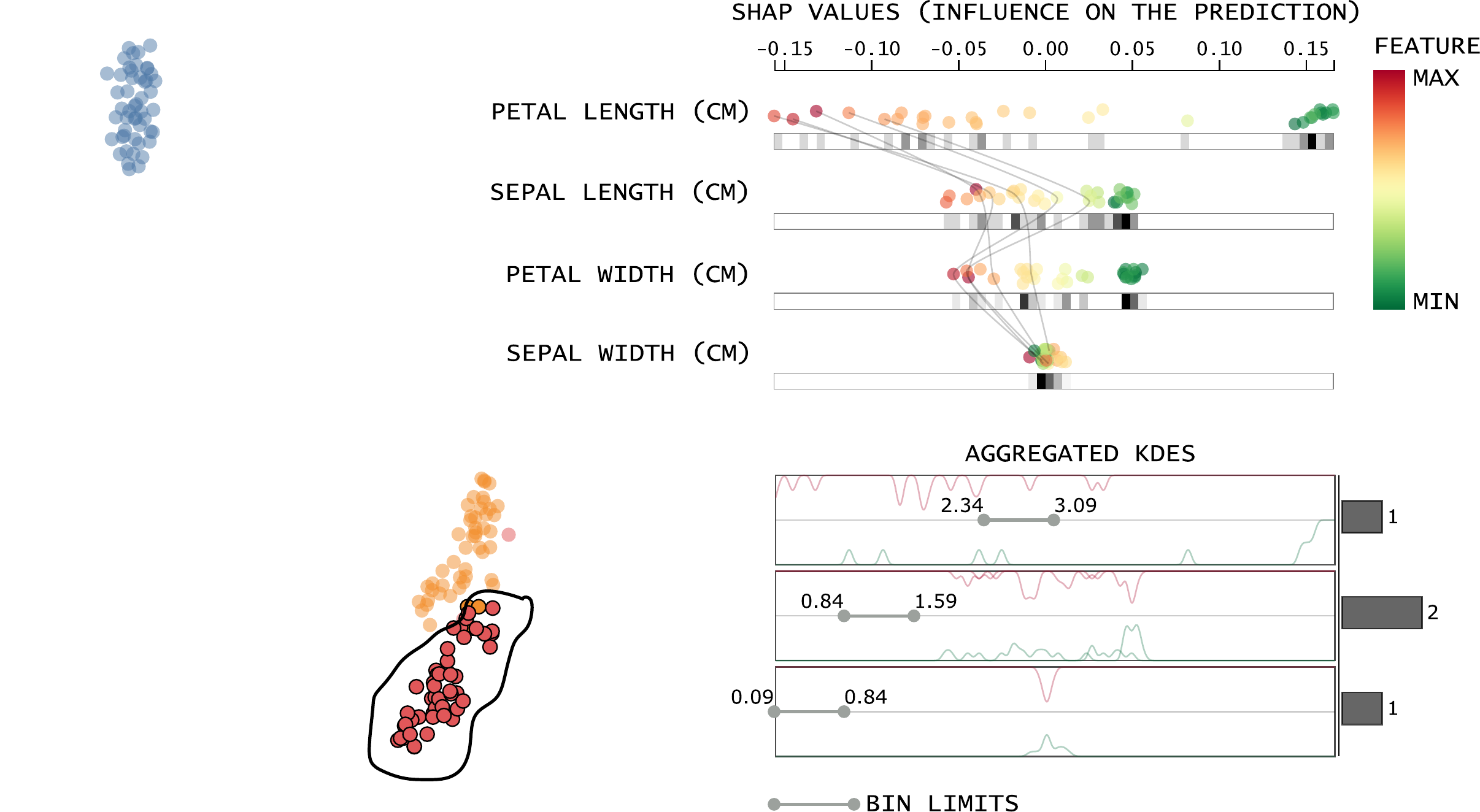}\label{fig:iris_dotplot-class2}}

    \caption{Two clusters with low separation. While the first three features show the same overall importance for these two clusters, \texttt{sepal width} plays a significant role in differentiating cluster orange (a).}
    \label{fig:iris_dotplot-class1-2}
\end{figure}

The inspection of clusters $1$ \img{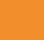} and $2$ \img{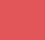} (see Figure~\ref{fig:iris_dotplot-class1-2}) presents similar \texttt{petal length} patterns (the most important feature). Higher values for such a feature are essential for cohesion (see negative Shapley values) for both clusters. Also, while the features of \texttt{sepal length} and \texttt{petal width} assume a different position on the importance ordering, their overall importance seems not to have much effect if we consider the density plot patterns. Finally, unlike clusters $0$ \img{figs/0.png} and $2$ \img{figs/2.png}, \texttt{sepal width} has a substantial influence on cluster $1$ \img{figs/1.png}, where lower values help characterize the clusters – see higher density for green areas of the dot plot in Figure~\ref{fig:iris_dotplot-class1}.

\section{Case Studies}
\label{sec:case-studies}

We start our analysis by focusing on medical datasets. We provide empirical evidence that Shapley values help to generate insights about pathologies and patients in different conditions. Then we analyze a dataset of quality indices in red wines, where we compare the characteristic found using Shapley values with their provided quality. Finally, we analyze a social dataset. 

All the projections were performed using \texttt{sklearn} implementation of t-SNE~\citep{Maaten_2008}, on a computer with the following configuration: Intel(R) Core(TM) i7-8700 CPU @ 3.20GHz, 32GB RAM, Windows 10 64 bits.

\subsection{Vertebral Column}

\begin{figure}[!htp]
\centering
\includegraphics[width=\linewidth]{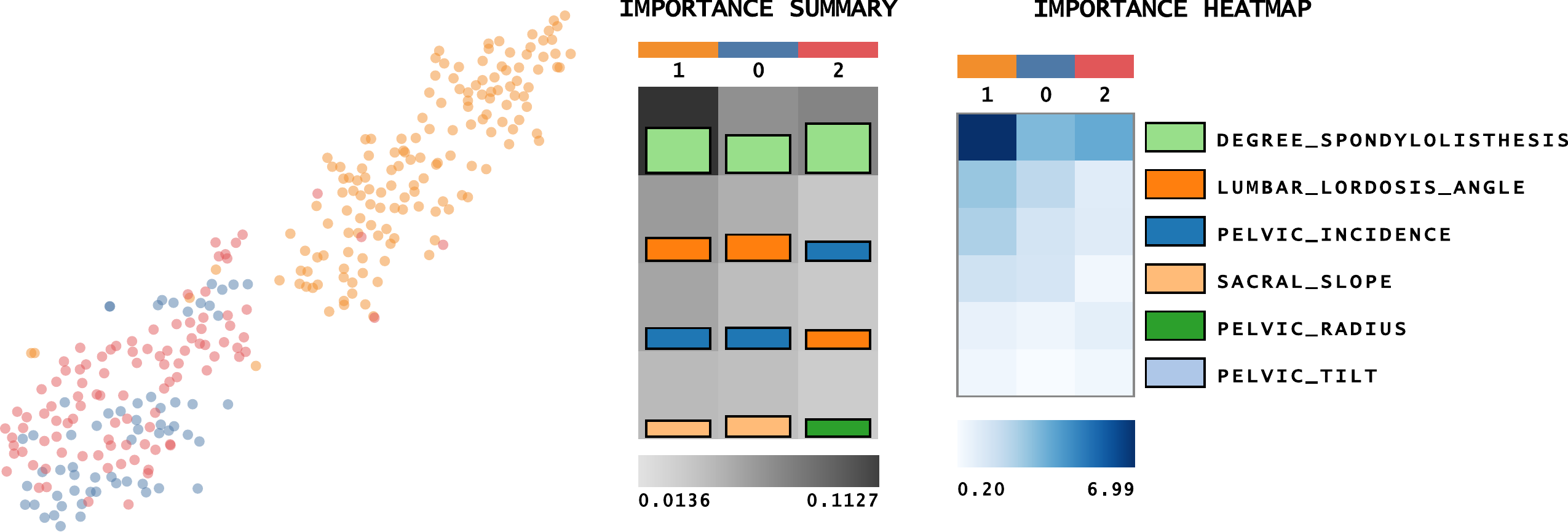}
\caption{Explanations for the \textit{Vertebral Column} dataset. There is a clear separation among patients with complications (Spondylolisthesis or Hernia) and the healthy ones. The \textbf{Importance Summary} also supports visualization of such characteristics.}\label{fig:vertebral-column_summary}
\end{figure}

In this first case study, we analyze a dataset containing six biomechanical features. The \textit{Vertebral Column} dataset~\citep{Dua2019} is composed by $310$ instances described by six features derived from the shape and orientation of the pelvis and lumbar spine: \texttt{pelvic incidence}, \texttt{pelvic tilt}, \texttt{lumbar lordosis angle}, \texttt{sacral slope}, \texttt{pelvic radius}, and \texttt{grade of spondylolisthesis}. Figure~\ref{fig:vertebral-column_summary} shows the projected instances, colored based on the ground truth classes: class $2$ \img{figs/2.png} for normal patients, class $0$ \img{figs/0.png} for patients with Hernia, and class $1$ \img{figs/1.png} for patients with Spondylolisthesis -- a disturbance of the spine in which a bone (vertebra) slides forward over the bone below it. There is a clearly separation of class $1$ \img{figs/1.png} among the remaining data points.

\begin{figure}[!htp]
\centering
\includegraphics[width=\linewidth]{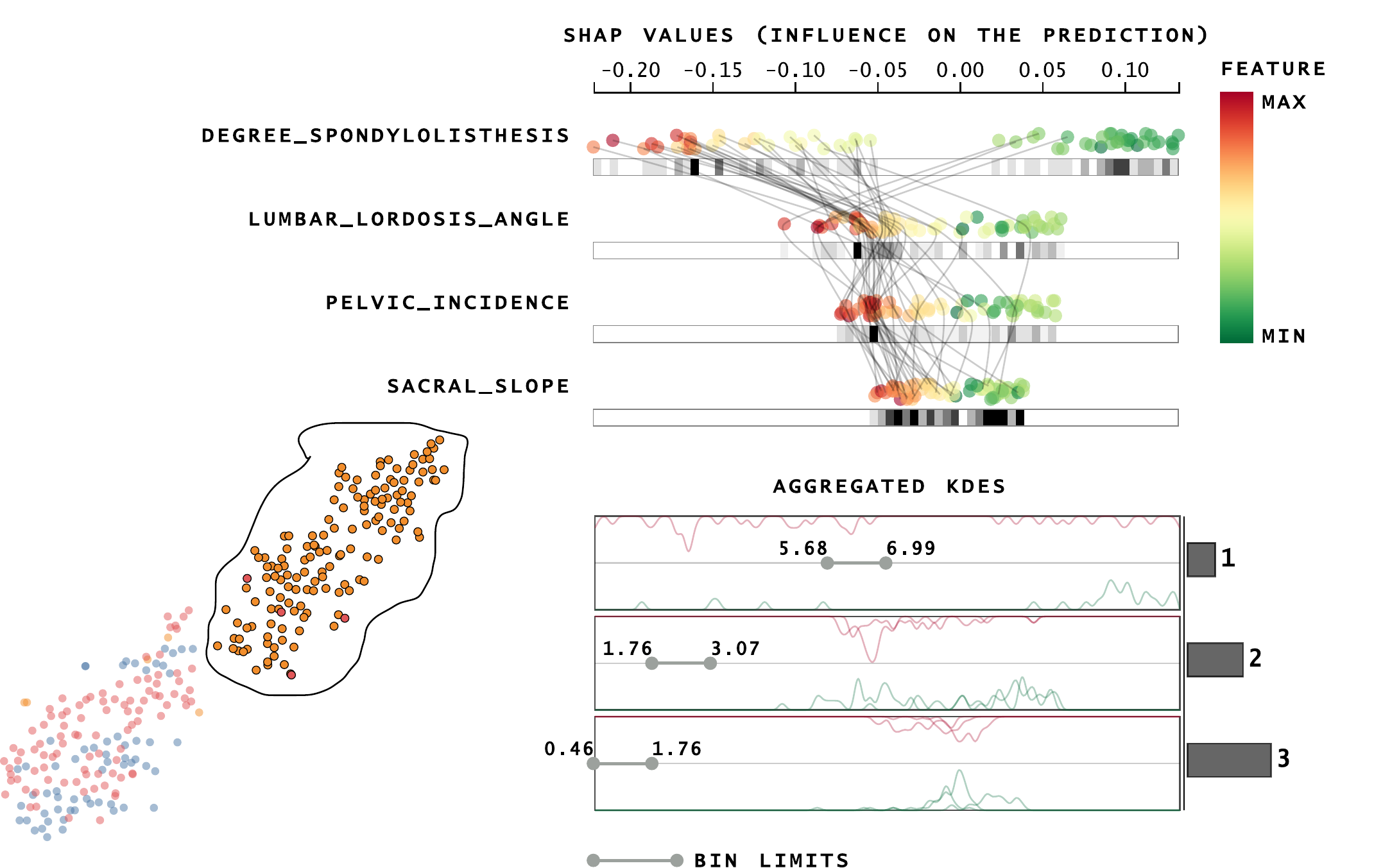}
\caption{Correlation between feature values and their importance for categorizing cluster $1$. The four most important features contributed to differentiate patients with Spondylolistyhesis from the others.}\label{fig:vertebral-column_dotplot-class1}
\end{figure}

The \textbf{Importance Summary} in Figure~\ref{fig:vertebral-column_summary} shows that the degree of \texttt{spondylolisthesis} is a determinant factor for the presence of Spondylolisthesis -- notice the feature legend next to the \textbf{Importance Heatmap}. Although it is important for characterizing all clusters, the color intensity and the heatmap values show that lower means of Shapley values can characterize the absence of Spondylolisthesis. By coordinating the scatter plot and the dot plot for class 1~\img{figs/1.png} (see Figure~\ref{fig:vertebral-column_dotplot-class1}, we visualize that this class's data points have higher values for the \texttt{degree of spondylolisthesis}. Further that, these higher feature values assume negative Shapley values and contribute to cluster formation. The other three most important features also contributed to clustering cohesion. The pelvic incidence angle measures the pelvic shape and determines the position of the sacral endplate~\citep{Tebet2013}. According to~\citet{Labelle2005}, pelvic incidence, sacral hill, pelvic tilt, and lumbar lordosis are greater in patients with developmental Spondylolisthesis.

\begin{figure}[!htp]
\centering
\includegraphics[width=\linewidth]{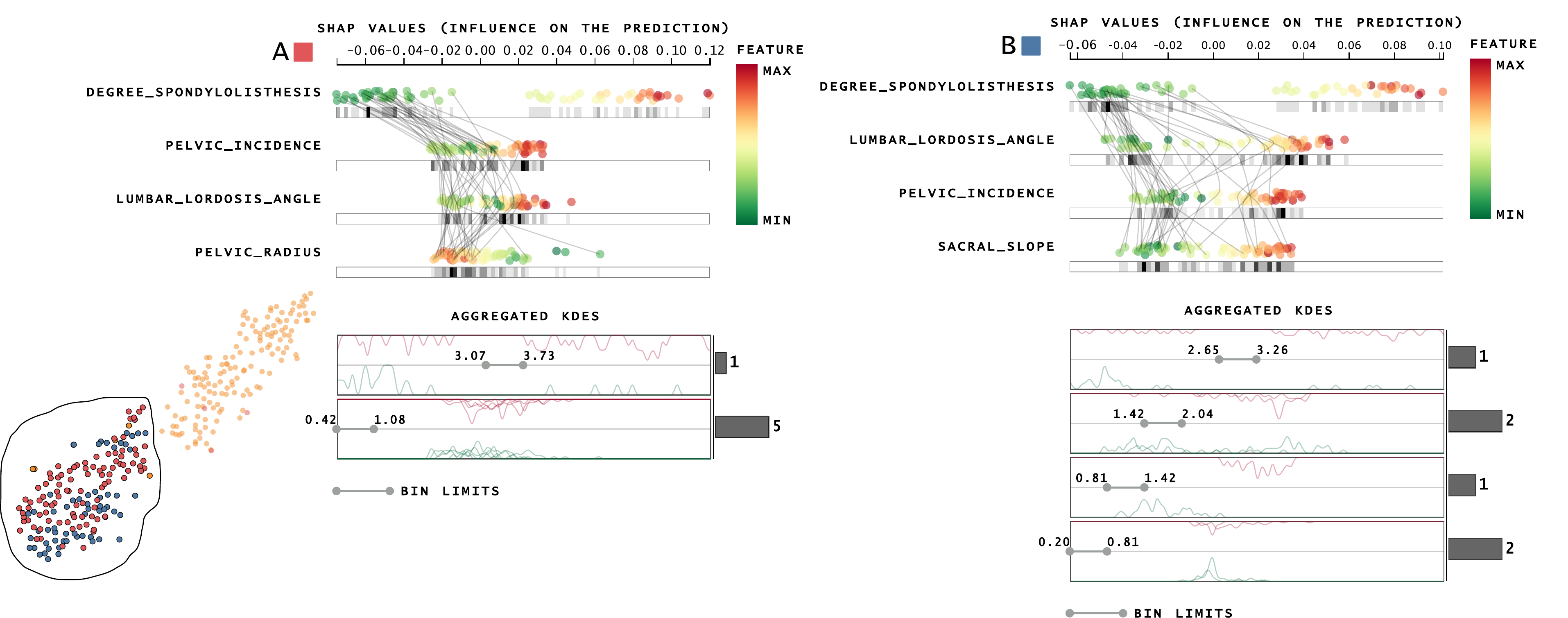}
\caption{Comparison between regular patients (A) and patients with Hernia (B). The difference in values for \texttt{sacral slope} was not good enough to impose a good separation on the dimensionality reduction result.}\label{fig:vertebral-column_dotplot-class0-2}
\end{figure}

Figure~\ref{fig:vertebral-column_dotplot-class0-2} shows the dot plots for classes $0$ \img{figs/0.png} and $2$ \img{figs/2.png}. Given the influence of the \texttt{degree of spondylolisthesis} for both classes, Spondylolisthesis's disturbance is not likely to be present in those instances due to the low feature values. The DR technique did not impose a clear separation between these classes because the features are not distinctive. However, there is a slight separation due to the degree of feature contribution -- as visualized in the aggregated KDEs. Finally, we see evidence for differentiating Hernia patients (B) from regular patients (A).  That is, patients with Hernia have low values for sacral slope (see this pattern for patients with Hernia~\img{figs/0.png} in Figure~\ref{fig:vertebral-column_dotplot-class0-2}(b)), which can indicate centralistic herniation~\citep{Roussouly2011}. 

\subsection{Indian Liver}
\label{sec:indian-liver}

\begin{figure}[!htp]
\centering
\includegraphics[width=\linewidth]{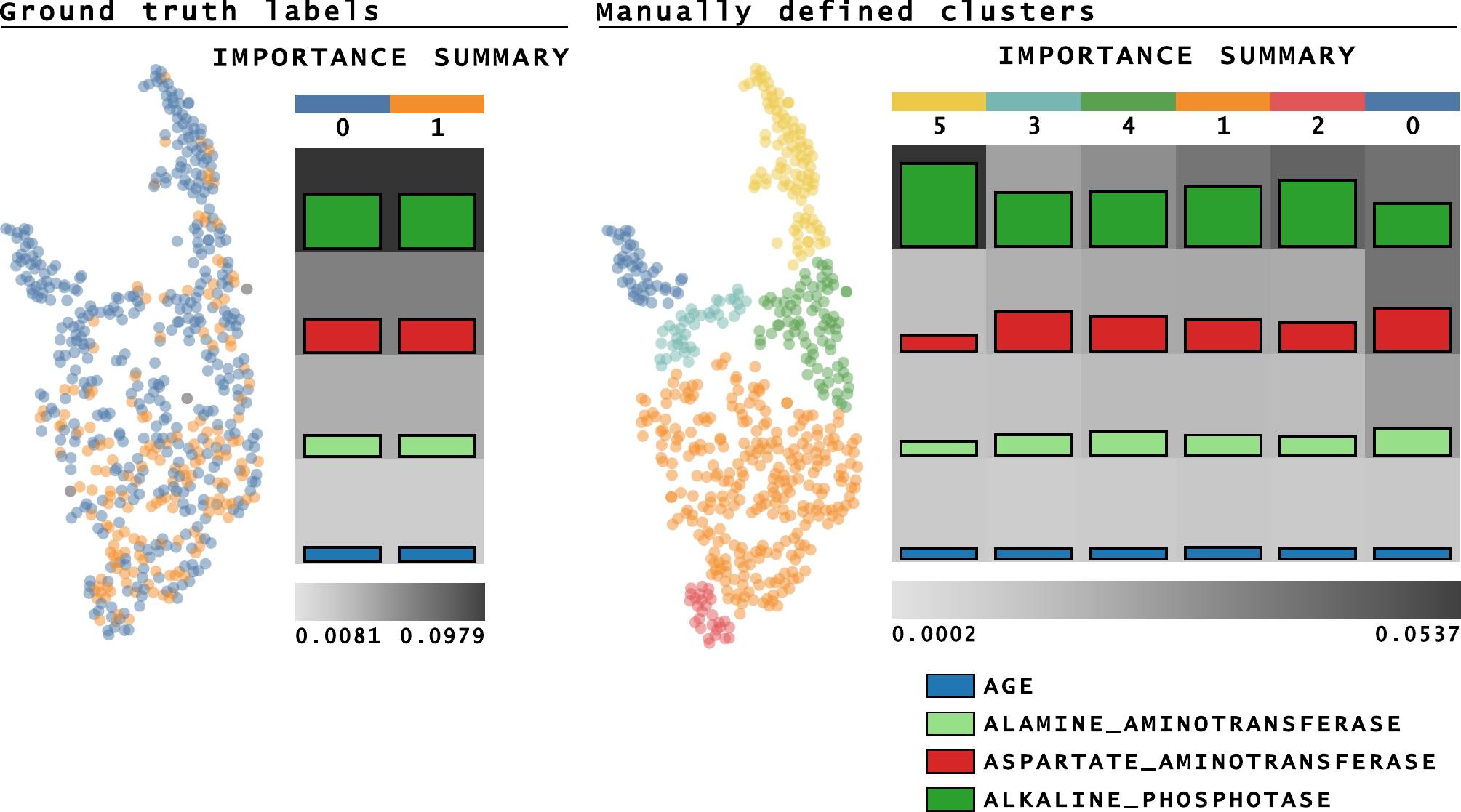}
\caption{Manually defined clusters to investigate the result of the dimensionality reduction process.}\label{fig:indian-liver_summary}
\end{figure}

In this case study, we investigate a dataset containing $416$ liver patient records and $167$ non-liver patient records. The instances are described by $10$ features: \texttt{Age}, \texttt{Gender}, \texttt{Total Bilirubin}, \texttt{Direct Bilirubin}, \texttt{Alkaline Phosphotase}, \texttt{Alamine Aminotransferase}, \texttt{Aspartate Aminotransferase}, \texttt{Total Proteins}, \texttt{Albumin}, and \texttt{Albumin and Globulin Ratio}. Coloring the projection based on the ground truth classes, we get nearly no distinction among the features' importance, where $0$ \img{figs/0.png} encodes patients with liver disease and $1$ \img{figs/1.png} encodes patients without liver disease. This characteristic is due to the similarity of data points– see how the two classes' instances are projected near to each other on the visual space in Figure~\ref{fig:indian-liver_summary}.

Since our tool allows for the manually definition of clusters, we inspect the clusters imposed by the DR technique. Figure~\ref{fig:indian-liver_summary} shows the clusters defined for this study case and respective \textbf{Importance Summary}. We used the following criteria for defining the clusters: select clusters with a majority of only one class (clusters $0$ \img{figs/0.png}, $5$ \img{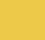}, and $3$ \img{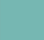}); select clusters with data points of mixed classes (clusters $4$ \img{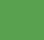} and $1$ \img{figs/1.png}); and select well-defined clusters (cluster $2$ \img{figs/2.png}). All clusters have the same important features 'and different contributions to the overall importance. Secondly, each cluster's three most determinant features consists of \texttt{Alamine Aminotransferase}, \texttt{Aspartate Aminotransferase}, and \texttt{Alkaline Phosphotase}, the latter playing a major role in categorizing cluster $5$ \img{figs/5.png}. Lastly, as a pattern noticed in the projection of Figure~\ref{fig:indian-liver_summary} (Ground truth labels), we can see that clusters $0$ \img{figs/0.png}, $3$ \img{figs/3.png}, and $5$ \img{figs/5.png} are the most distinctive.

\begin{figure}[!htp]
\centering
\includegraphics[width=\linewidth]{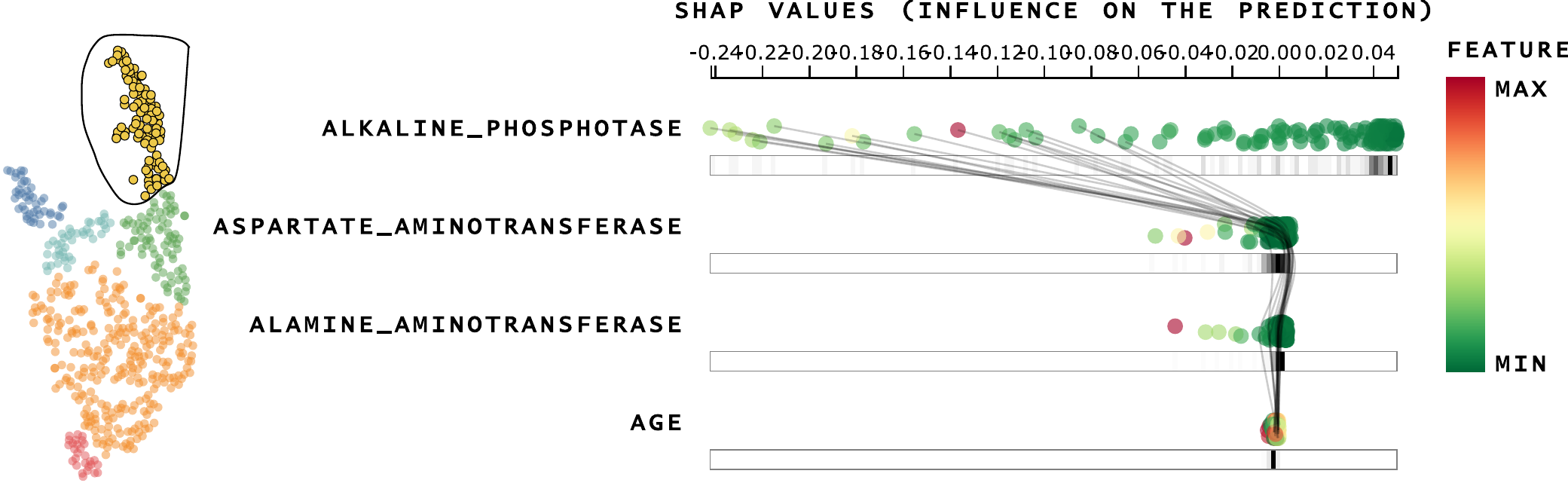}
\caption{Alkaline phosphotase's high feature values can indicate cholestatic liver disease, obstruction of bile ducts, or liver cancer. Nevertheless, young and elderly can also present higher values of \texttt{Alkaline Phosphotase}. }\label{fig:indian-liver_dotplot-class5}
\end{figure}

Assessing the dot plot of cluster $5$ \img{figs/5.png} (see Figure~\ref{fig:indian-liver_dotplot-class5}) and recalling that such cluster has a majority of patients with liver disease, there is an evident influence of patients with higher \texttt{Alkaline Phosphotase}. According to~\citet{Lowe2018}, high values of Alkaline Phosphotase lies in the diagnosis of cholestatic liver disease~\citep{Dhillon2012}. Alkaline Phosphotase can also present higher values when the bile ducts are blocked or by liver cancer~\citep{ALP}. Finally, readers may ask why a few patients with no liver pathology have high \texttt{Alkaline Phosphotase} as well. This happens because the levels increase for young people and the elderly~\citep{Lowe2018} (our case).

\begin{figure}[!htp]
\centering
\includegraphics[width=\linewidth]{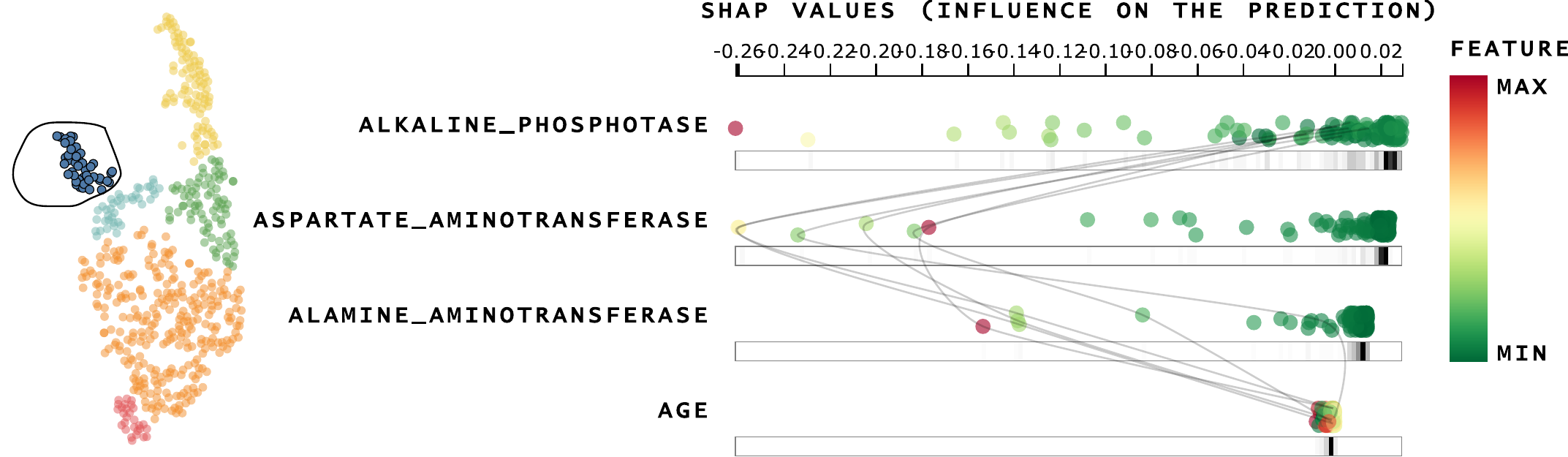}
\caption{Blue cluster could indicate patients with Hepatitis that usually present lower levels of Alamine Aminotransferase and higher levels of Aspartate Aminotransferase and Alamine Aminotransferase.}\label{fig:indian-liver_dotplot-class0}
\end{figure}

Another interesting pattern appears when analyzing the dot plot of cluster $0$ \img{figs/0.png} in Figure~\ref{fig:indian-liver_dotplot-class0}. We can note that while almost every feature contributed to the no cohesion of the clusters -- see the density of points around $0.2$ for \texttt{Alkaline Phosphotase}, around $0$ for \texttt{Age} and the aggregated KDE's -- the high values of features \texttt{Aspartate} and \texttt{Alamine} \texttt{Aminotransferase} contributed for the distinction of the clusters. Accordingly, since all the instances of cluster $0$ \img{figs/0.png} have liver disease, the two features are likely to influence the pathology. While lower levels of \texttt{Aspartate} and \texttt{Alamine} \texttt{Aminotransferase} are expected, higher levels of these two features indicate liver diseases, such as viral infection or acute Hepatitis~\citep{Dhillon2012,Anadon2019}. For this set of features, all of the patients present liver disease and lower \texttt{Alkaline Phosphotase} levels. Although we would need a more detailed dataset to confirm our hypothesis, these instances could be patients with Hepatitis, where Alkaline Phosphotase is usually much less elevated than Aspartate and Alamine Aminotransferase.

\begin{figure}[!htb]
    \centering
    \subfloat[The features did not have much influence on categorizing cluster $1$.]{\includegraphics[width=\linewidth]{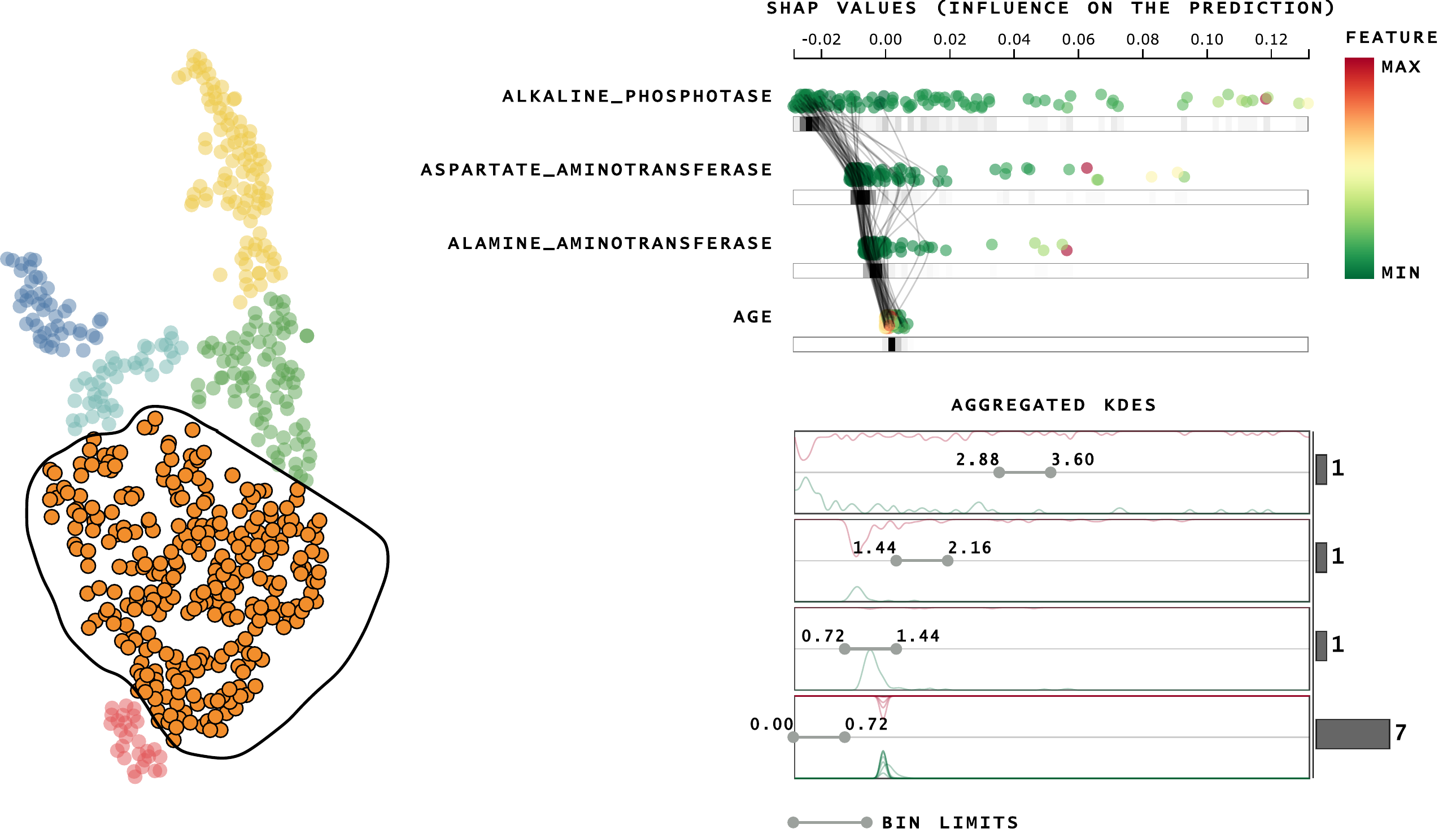}\label{fig:indian-liver_dotplot-class1}}   

    \subfloat[Strong influence from \texttt{Alkaline Phosphotase}.]{\includegraphics[width=\linewidth]{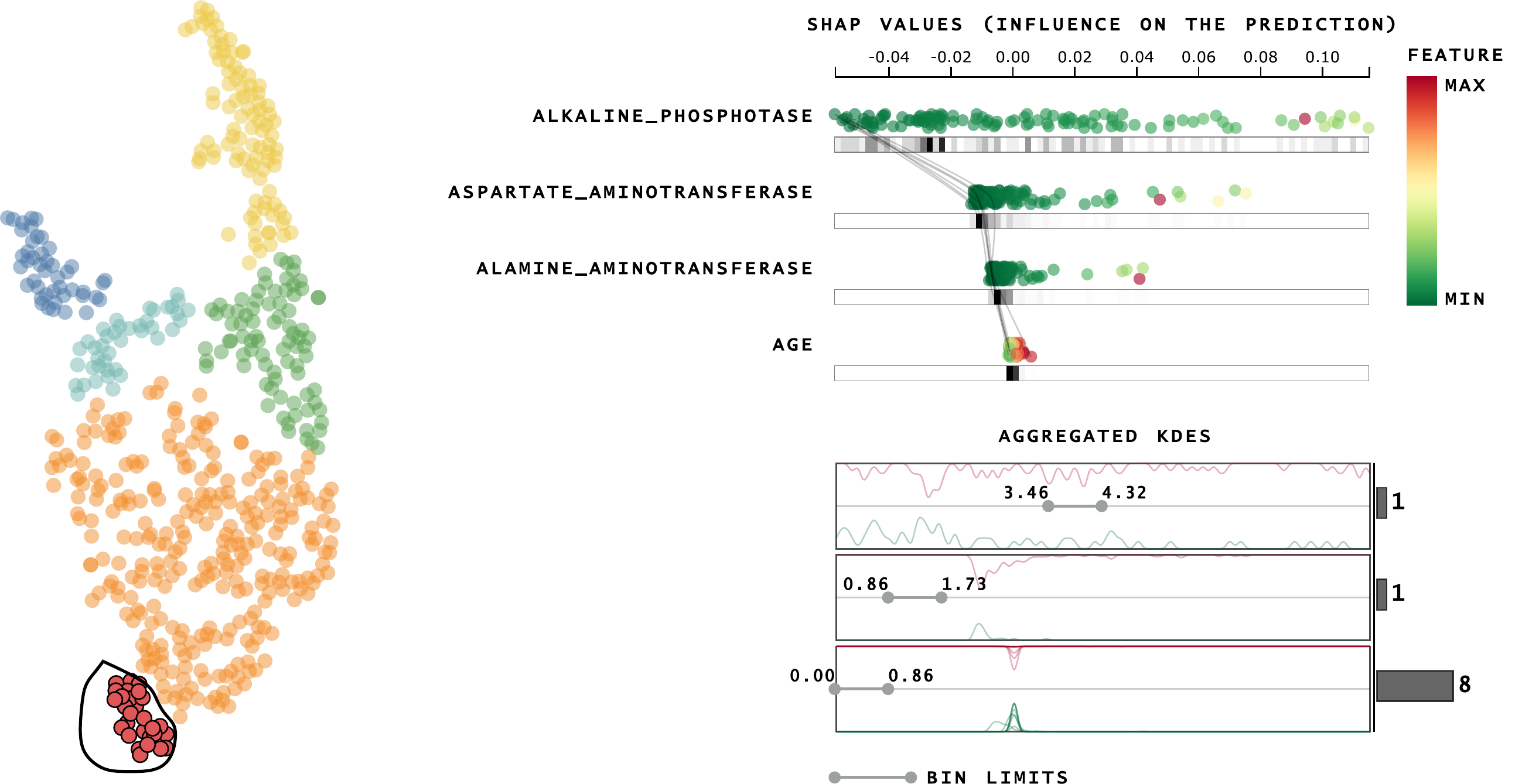}\label{fig:indian-liver_dotplot-class2}}

    \caption{Comparison between clusters $1$ and $2$. These two clusters present different patterns for \texttt{Age} and the remaining of the features as seen in the \textbf{Aggregated KDEs}.}
    \label{fig:indian-liver_dotplot-class1-2}
\end{figure}

Figure~\ref{fig:indian-liver_dotplot-class1-2} shows the dot plots for classes $1$ \img{figs/1.png} and $2$ \img{figs/2.png}. Notice that these classes present lower feature values for the three most important features, contributing to the spatial distance from the well-defined clusters where most patients have liver disease. Moreover, note how these clusters differ only on the patients' age, which helped determine a separation among the two clusters.

\begin{figure}[!htp]
\centering
\includegraphics[width=\linewidth]{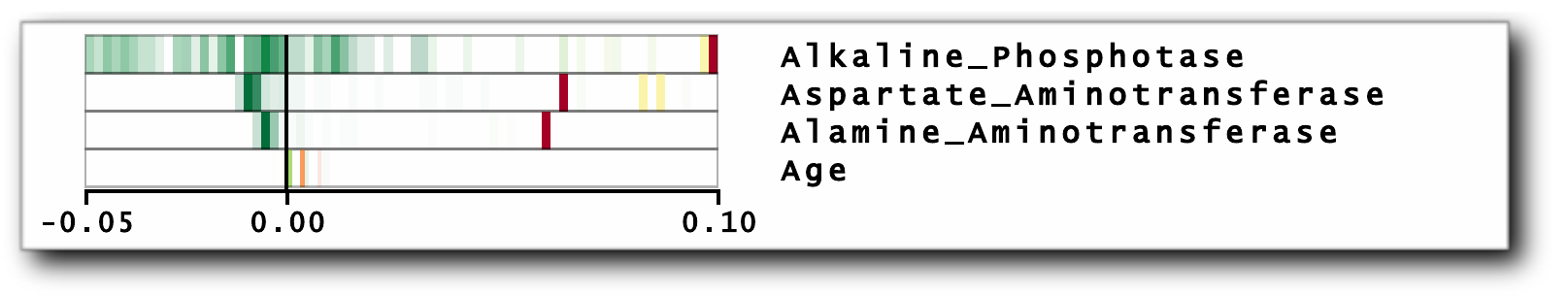}
\caption{Summary representation of features values for cluster $4$ reveals a group of patients with liver disease. Lower values for the first three features can be related to chronic hepatitis, obstruction of bile ducts, or cirrhosis.}\label{fig:indian-liver_summary-cluster4}
\end{figure}

The question is why several instances of patients with the disease have low values for features \texttt{Alk. Phosphotase}, \texttt{Asp. Aminotransferase}, and \texttt{Alam. Aminotransferase}. There are two possible answers: first, the feature space may not be enough to describe and impose a separation between the two classes; second, levels of Aspartate Aminotransferase and Alamine Aminotransferase do not present high levels for some liver disease, such as chronic hepatitis, obstruction of bile ducts, or cirrhosis~\citep{ALP,ALT}. Cluster $4$ \img{figs/4.png}, for example, has more patients with liver disease. The summary of Shapley values in Figure~\ref{fig:indian-liver_summary-cluster4} shows much lower values of the three most important features. In other words, by having negative Shapley values (which help at characterizing the cluster) for the features with lower values, we hypothesize that this cluster corresponds to patients with a liver pathology due to the lower values of the \texttt{Alkaline Phosphotase}, \texttt{Aspartate}, and \texttt{Alamine Aminotransferase} features.

\iffalse
\begin{figure}[!htp]
\centering
\includegraphics[width=\linewidth]{figs/indian-dotplot-class4.png}
\caption{.}\label{fig:indian-liver_dotplot-class4}
\end{figure}
\fi

\subsection{Red Wine Quality}

\begin{figure}[!htp]
\centering
\includegraphics[width=\linewidth]{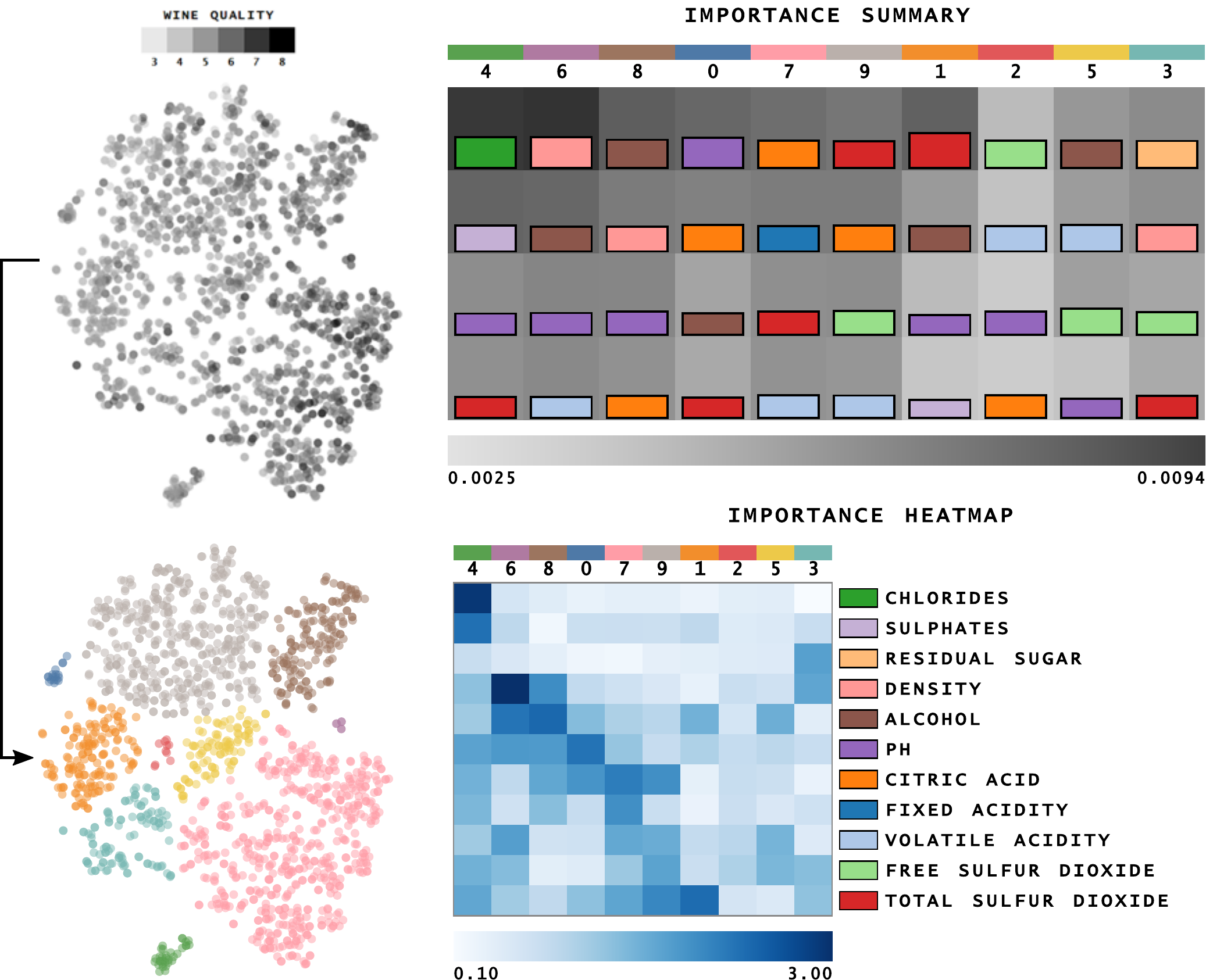}
\caption{Dimensionality reduction results for the \textit{Red Wine Quality} dataset colored using intensity to encode quality index and using colors to depict clusters. }\label{fig:wine-quality_summary}
\end{figure}

For this case study, we understand the quality of red wines \citep{Dua2019} by explaining manually defined clusters after dimensionality reduction. Figure~\ref{fig:wine-quality_summary} shows a t-SNE projection with grayscale encoding wine quality. The dataset contains $1599$ instances described by $11$ features: \texttt{fixed acidity}, \texttt{volatile acidity}, \texttt{citric acid}, \texttt{residual sugar}, \texttt{chlorides}, \texttt{free sulfur dioxide}, \texttt{total sulfur dioxide}, \texttt{density}, \texttt{pH}, \texttt{sulphates}, and \texttt{alcohol}.

Unlike the previous case studies, ClusterShapley reveals a more heterogeneous set of feature importance for this dataset. However, certain similarities can be noticed, such as clusters $6$ \img{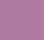} and $8$ \img{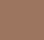} and clusters $2$ \img{figs/2.png} and $5$ \img{figs/5.png}. The most cohesive cluster ($4$ \img{figs/4.png}) -- more spatially distant from the others -- has a few features that contributed to its distinction: \texttt{chlorides} and \texttt{sulphates}. There are other characteristics. For example, the \texttt{pH} index was determinant for characterizing cluster $0$ \img{figs/0.png}, \texttt{density} in cluster $6$ \img{figs/6.png}, and for cluster $2$ \img{figs/2.png}, the features contributed equally.

\begin{figure}[!htp]
\centering
\includegraphics[width=\linewidth]{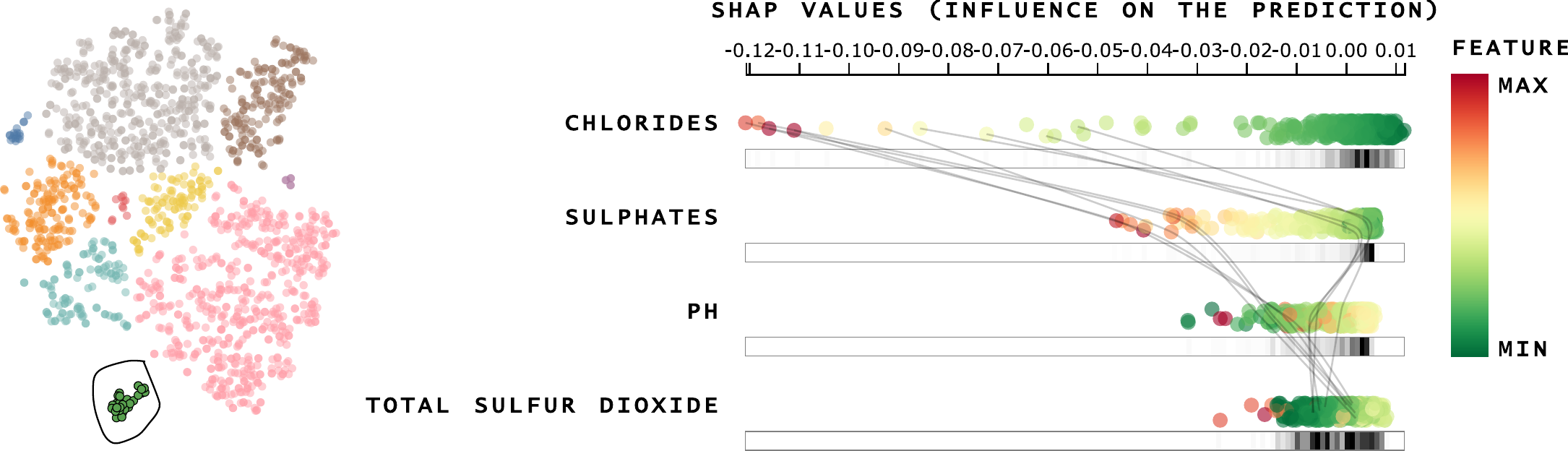}
\caption{ Cluster $4$ is well defined and distinguishable due to the salty flavor (high levels of \texttt{chrolides}) of the wines.}\label{fig:wine-quality_dotplot-class4}
\end{figure}

Now let us inspect cluster $4$ \img{figs/4.png} to assess the contribution of \texttt{chlorides} and \texttt{sulphates}. Figure~\ref{fig:wine-quality_dotplot-class4} shows the line segments and Shapley values for class $4$ \img{figs/4.png}. Notice that while other clusters present lower \texttt{chloride} values and \texttt{sulphates}, cluster $4$ \img{figs/4.png} presents higher values for these features. Compared with the other Shapley values, \texttt{chlorides} and \texttt{sulphates} contributed a lot to the cluster separation, i.e., Shapley values' negative contribution. While \texttt{chlorides} measure salt in the wine, \texttt{sulphates} contribute to sulfur dioxide gas levels and act as an antimicrobial and antioxidant~\citep{Cortez2009}. Thus, such a cluster describes salty wines, being very distinguishable from the others.

\begin{figure}[!htb]
    \centering
    \subfloat[Lower levels of \texttt{Alcohol} were determinant for the lower quality of the wines in cluster $0$.]{\includegraphics[width=\linewidth]{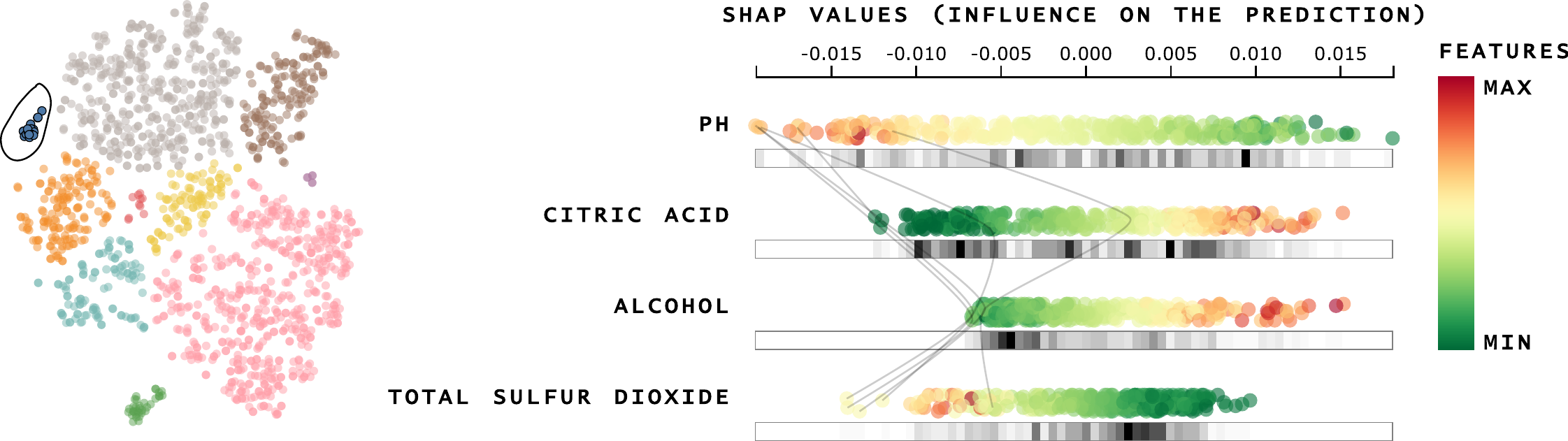}\label{fig:wine-quality_dotplot-class0}}       

    \subfloat[Higher levels of \texttt{Alcohol} determine higher quality for wines in cluster $8$.]{\includegraphics[width=\linewidth]{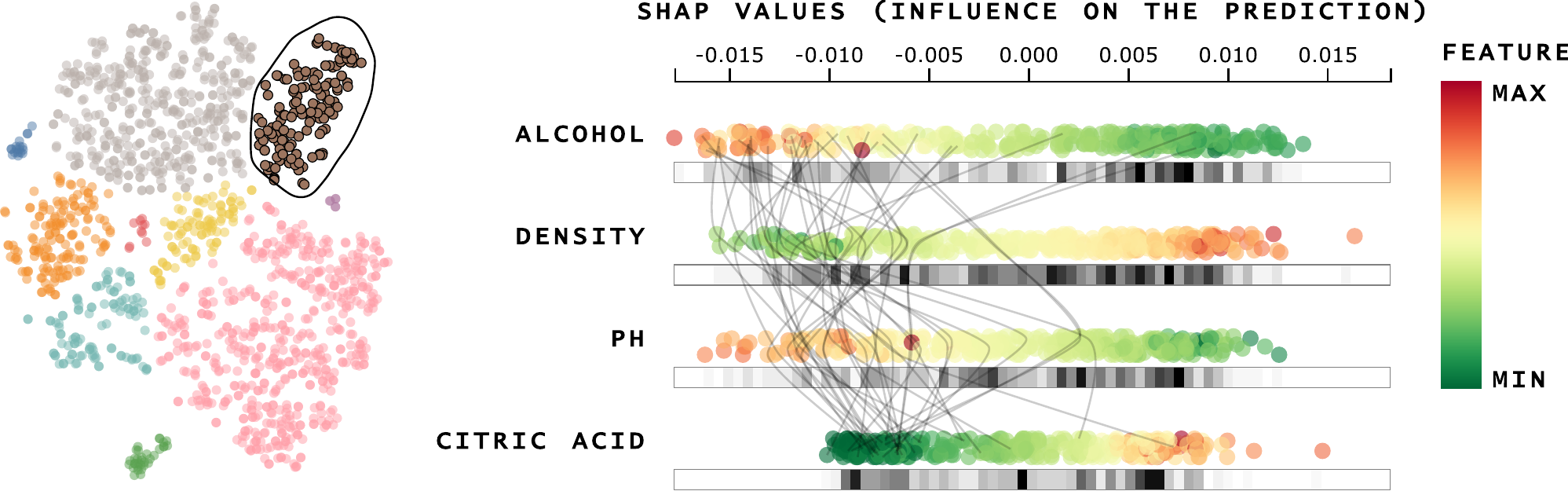}\label{fig:wine-quality_dotplot-class8}}
    
    \caption{Comparison between two clusters projected distant which share two important features. The difference between alcohol levels was determinant for differentiating the quality of the wine in these two clusters.}
    \label{fig:wine-quality_dotplot-class0-8}
\end{figure}

Figure~\ref{fig:wine-quality_dotplot-class0-8} shows the dot plot of clusters $0$ \img{figs/0.png} and $8$ \img{figs/8.png}. Both of the clusters influence higher values of \texttt{pH} index. However, the two most influential features of cluster $0$ \img{figs/0.png} are only the third and fourth in the ordering of cluster $8$ \img{figs/8.png}. Moreover, the two groups of instances benefit from higher \texttt{pH} values and lower values of \texttt{critic ACID}; cluster $8$ \img{figs/8.png} presents higher values for \texttt{alcohol}, configuring among the more alcoholic wines and more tasteful according to the quality index.

Recalling the DR result encoding the quality index, we now inspect cluster $7$ \img{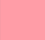} with higher quality wines (higher intensity in Figure~\ref{fig:wine-quality_summary}). Cluster $7$ \img{figs/7.png} presents \texttt{Fixed Acidity} as the second most important feature. A combination of higher values \texttt{Fixed Acidity} and lower \texttt{Volatile Acidity} values means higher quality for the wines. For instance, volatile acid in higher concentrations results in an unpleasant aroma and taste~\citep{Ezekeil2004}. Another characteristic that explains the concentration of higher quality is the higher levels of \texttt{Citric Acid}, which contributes to the wine's freshness. Besides that, lower values for \texttt{Total Sulfur Dioxide} feature also contribute to the higher quality, i.e., when in lower values, \texttt{Total Sulfur Dioxide} do not add flavor or smell to the wine~\citep{Cortez2009}.

\begin{figure}[h!]
\centering
\includegraphics[width=\linewidth]{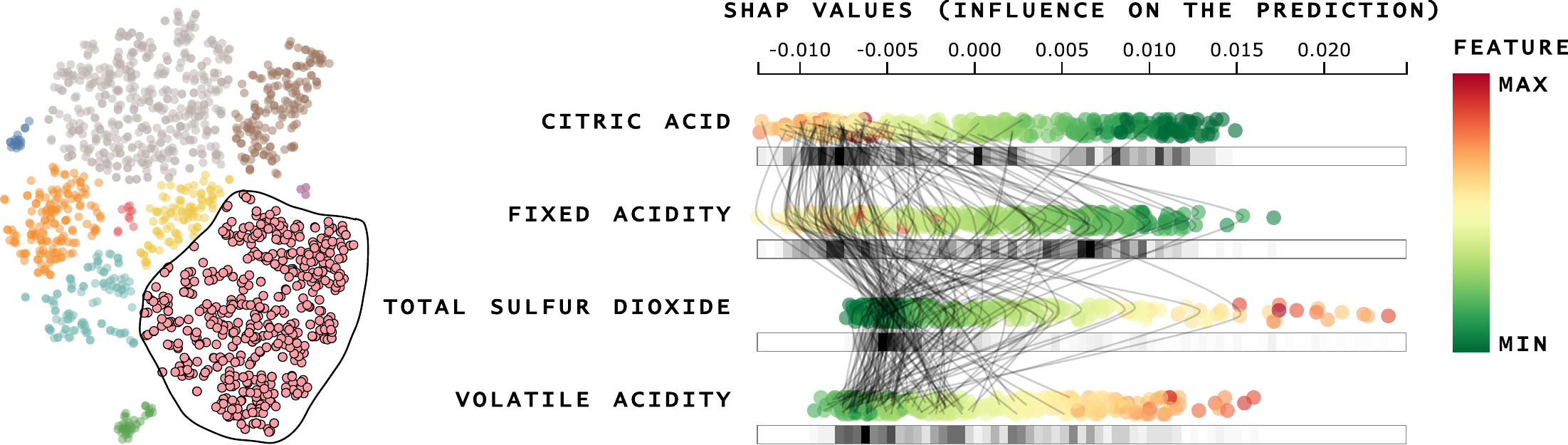}
\caption{A combination of higher values \texttt{Fixed Acidity} and lower values of \texttt{Volatile Acidity} can be unpleasant at higher levels and can explain higher quality for the data instances.}\label{fig:wine-quality_dotplot-class7}
\end{figure}

\begin{figure}[!htp]
\centering
\includegraphics[width=\linewidth]{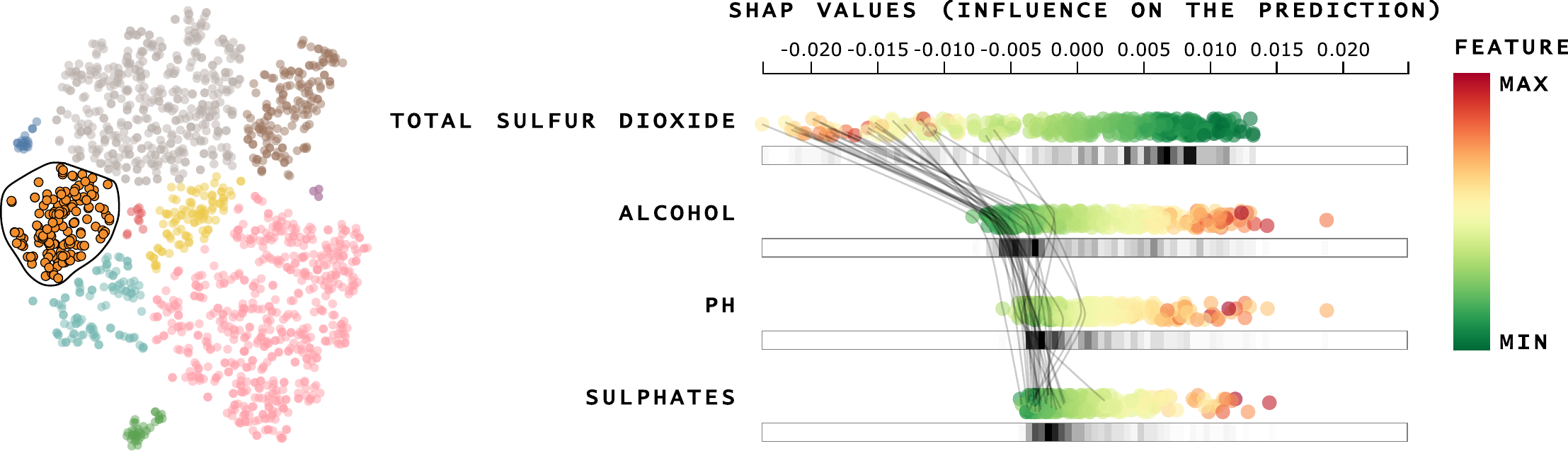}
\caption{At higher levels, \texttt{Total Sulfur Dioxide} tends to add flavor to the wine and be noticed in the nose. Such an important feature is the meaning of higher concentration of low-quality wines in cluster $1$.}\label{fig:wine-quality_dotplot-class1}
\end{figure}

Finally, focusing on cluster $1$ \img{figs/1.png}, Figure~\ref{fig:wine-quality_dotplot-class1} shows that \texttt{Total Sulfur Dioxide} (SO2) is a determinant for the concentration of lower quality wines. While lower concentrations of S02 are mostly undetectable, higher concentrations of S02 becomes evident in the nose and taste. The other most important feature contributes to the quality as well. As we could see for clusters $4$ \img{figs/4.png} and $8$ \img{figs/8.png}, lower values for \texttt{Alcohol} were also responsible for lower quality in cluster $4$ \img{figs/4.png}.

\subsection{Communities and Crime}
\label{sec:communities-crime}

\begin{figure}[!htp]
\centering
\includegraphics[width=\linewidth]{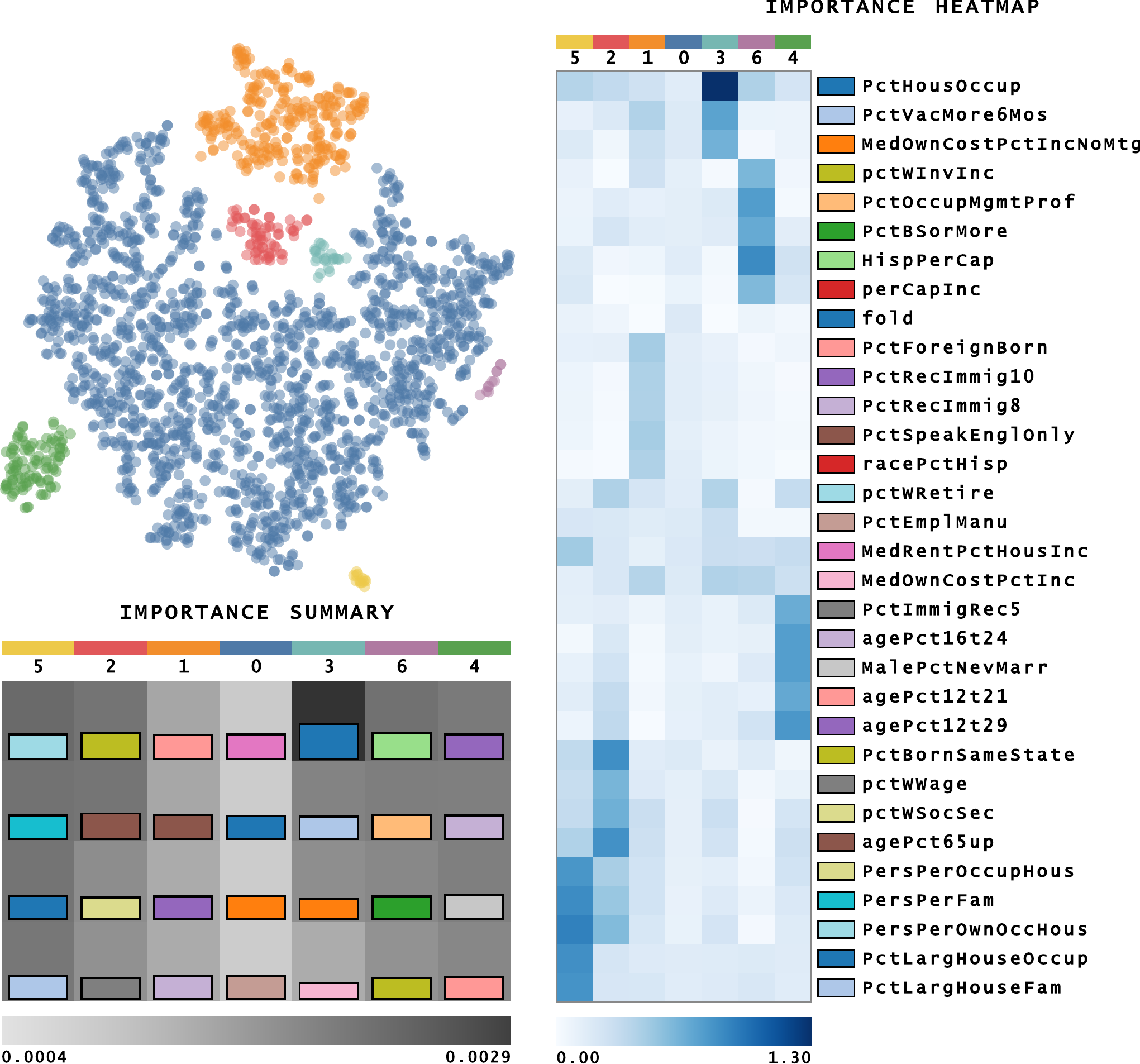}
\caption{Feature contributions and the dimensionality reduction result for the \textit{Communities and Crime} dataset.}\label{fig:communities-crime_summary}
\end{figure}

In this final case study, we explore the Communities and Crime dataset~\citep{Redmon2002,Dua2019}. We used the dataset preprocessed by~\citet{Fujiwara2019} to compare the patterns returned by our approach. The dataset contains $2215$ instances described by $128$ features. Figure~\ref{fig:communities-crime_summary} shows the projection result with clusters manually selected together with the overview visualization of feature importance.

\begin{figure}[!htp]
\centering
\includegraphics[width=\linewidth]{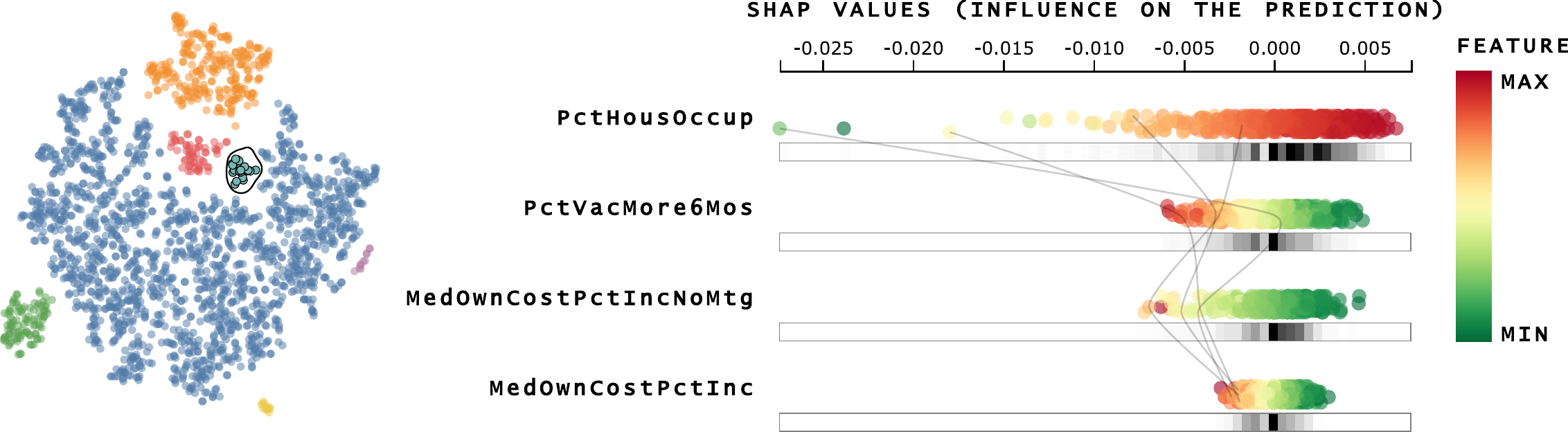}
\caption{Such community is very distinguishable from the other for having a lower percentage of houses occupied \texttt{PctHousOccup}.}\label{fig:communities-crime_dotplot-class3}
\end{figure}

Apart from clusters $0$ \img{figs/0.png} and $3$ \img{figs/3.png}, the most important features are very heterogeneous among the clusters. The first thing to notice is how \texttt{PctHousOccup} (percent of occupied houses) is important for characterizing cluster $3$ \img{figs/3.png}. Such cluster is different from the others by having a lower percentage of houses occupied -- as shown in the dot plot of Figure~\ref{fig:communities-crime_dotplot-class3} -- indicating communities in safer areas.

\begin{figure}[!htb]
    \centering
    \subfloat[A community that has higher percent of males who have never married (\texttt{MalePctNevMarr}) due to the high percent of young people (\texttt{AgePct12T29}, \texttt{AgePct16T24}, and \texttt{AgePct12T21}).]{\includegraphics[width=\linewidth]{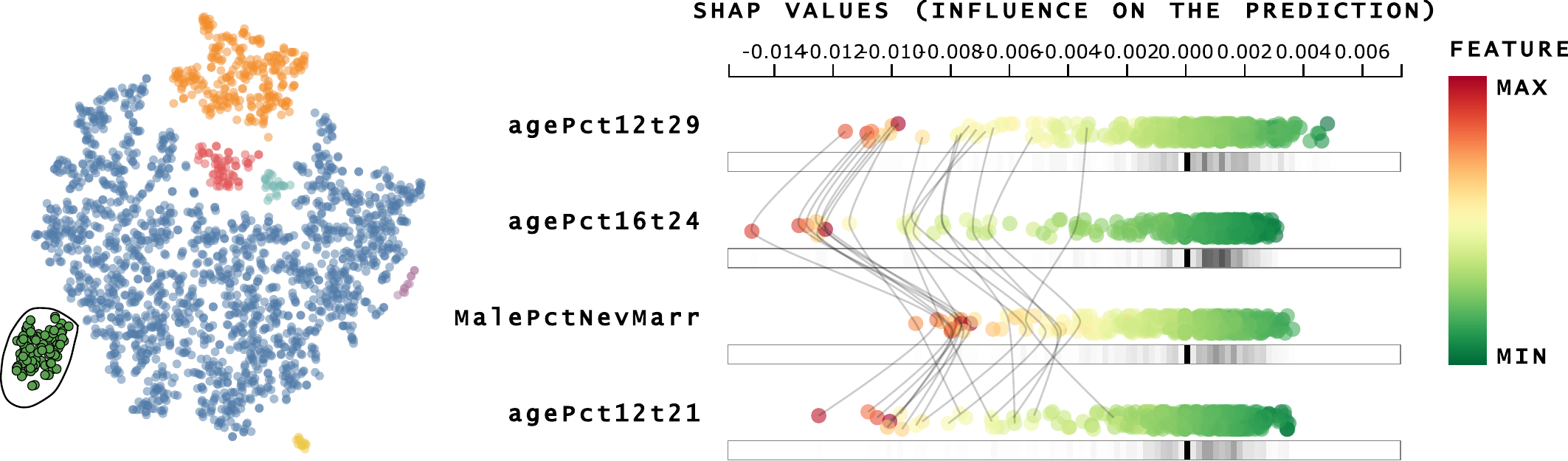}\label{fig:communities-crime_dotplot-class4}}      

    \subfloat[A community that has bigger families/houses (\texttt{PctLargHouseFam}) and suffer from robbery (\texttt{PctLargHouseOccup}).]{\includegraphics[width=\linewidth]{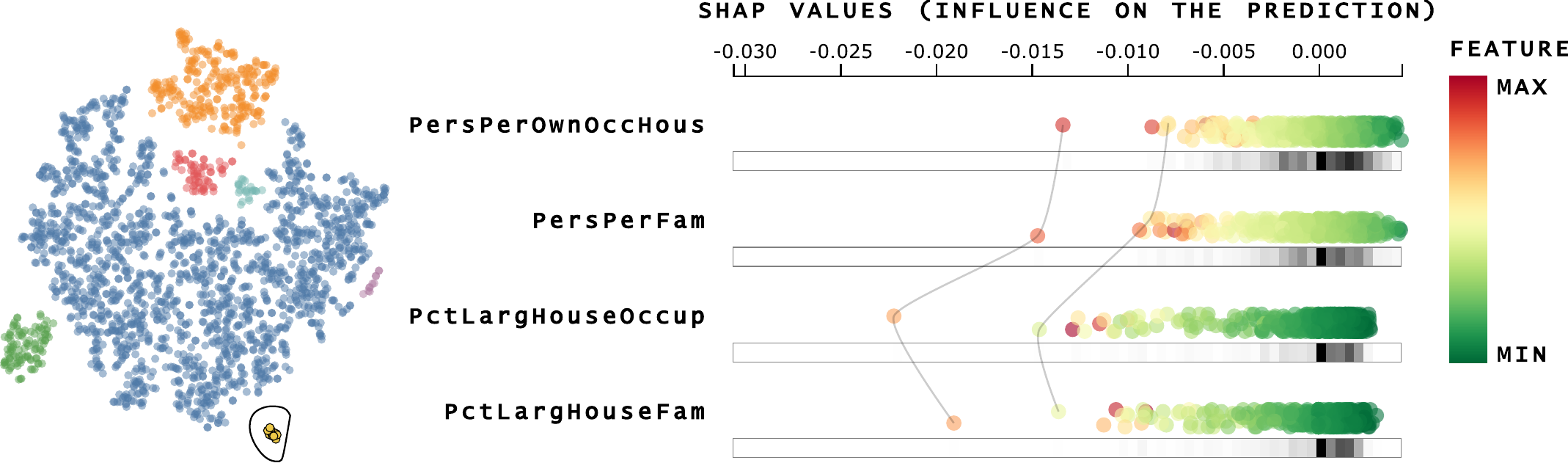}\label{fig:communities-crime_dotplot-class5}}

    \caption{Detailed analysis of clusters $4$ and $5$.}
    \label{fig:communities-crime_dotplot-class4-5}
\end{figure}

Another interesting pattern is the importance of the features \texttt{PctLargHouseFam} (percent of large family households) and \texttt{PctLargHouseOccup} (percent of large house occupied) for cluster $5$ \img{figs/5.png} (see Figure~\ref{fig:communities-crime_dotplot-class4}), as well as the features \texttt{AgePct12T29}, \texttt{AgePct16T24} (percent of population between $12$ and $29$, and percent of population between $16$ and $24$, respectively), and for the feature \texttt{MalePctNevMarr} (percent of Males who have never married) for cluster $4$ \img{figs/4.png} (see Figure~\ref{fig:communities-crime_dotplot-class5}). For cluster $5$ \img{figs/5.png}, these features' higher values contribute to cohesion, representing bigger families and houses. The same pattern happens to cluster $4$ \img{figs/4.png}, the higher percentage of age explains the higher percentage of males that have never married. Since the age range in these features is somewhat low, the percentage of males who have never married tends to be high due to cultural aspects.

Finally, cluster $1$ \img{figs/1.png} corresponds to a group of immigrants. The four most important features are \texttt{PctForeignBorn} (percent of people born in another country), \texttt{PctRecImmig8}, \texttt{PctRecImmig10} (percent of the population who have immigrated within the last 8 and 10 years), and \texttt{PctSpeakEngOnly} (percent of people who speak only English). While the first three features present high values, the latter shows low values. People who have immigrated are more likely to speak another language. Such analysis exemplifies the semantic power of our approach.

\begin{figure}[!htp]
\centering
\includegraphics[width=\linewidth]{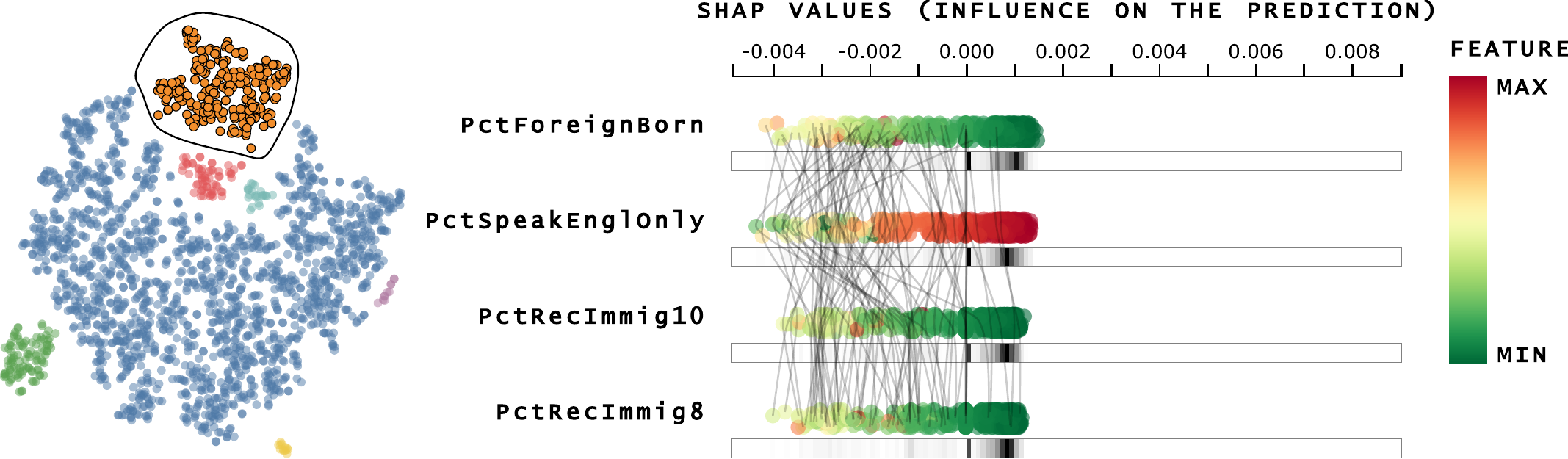}
\caption{A community of immigrants due to the high percent of immigrants (\texttt{PctForeignBorn}, \texttt{PctRecImmig8}, and \texttt{PctRecImmig10}) who speak more than one language besides English (\texttt{PctSpeakEngOnly}).}\label{fig:communities-crime_dotplot-class1}
\end{figure}

\subsection{Using clustering techniques}

\begin{figure*}[!htp]
\centering
\includegraphics[width=\linewidth]{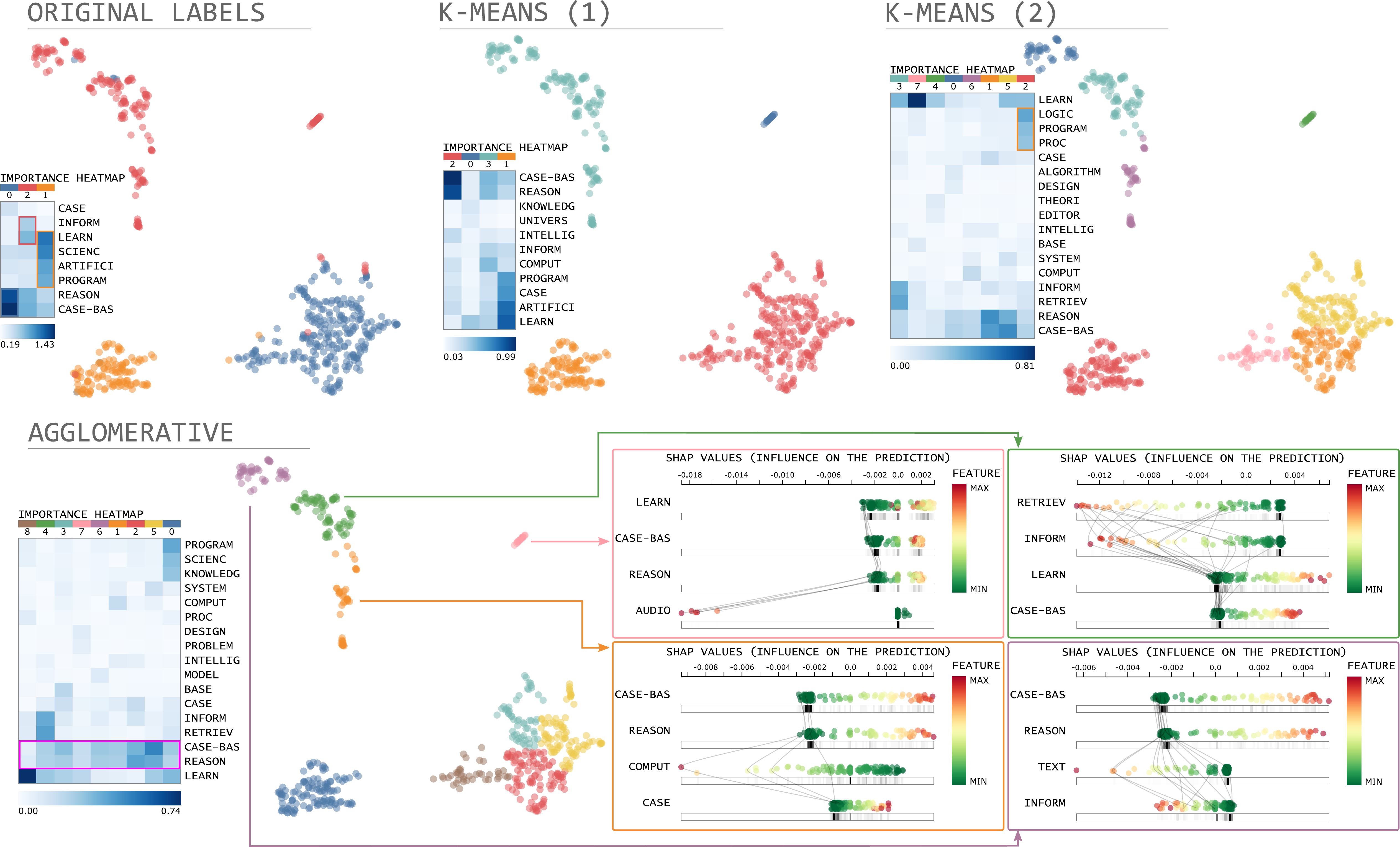}
\caption{Using clustering techniques to feed ClusterShapley with different clustering results. The original dataset (\texttt{original labels}) has three labels, and we use three clustering configurations: four clusters using $k$-Means (\texttt{k-means (1)}), eight clusters using $k$-Means (\texttt{k-means (2)}), and nine clusters using Agglomerative clustering (\texttt{agglomerative}).}\label{fig:clustering-analysis}
\end{figure*}

In this section, we aim to analyze ClusterShapley using clustering algorithms. To this end, we choose the CBR-ILP-IR~\citep{Paulovich_2008} document collection, also aiming to evaluate our technique for thousands of dimensions. The CBR-ILP-IR dataset contains $574$ documents representing papers of three fields: Case-based Reasoning (CBR), Intuitive Logic Programming (ILP), and Information Retrieval (IR). We followed~\citet{Eler2018} to preprocess the dataset, using Porter Stemming, removing terms with a frequency below one and \textit{tf-idf} transformation. The resulting dataset consists of $574$ papers per $18694$ terms. Figure~\ref{fig:clustering-analysis} shows the resulting clusters on a UMAP~\citep{McInnes2018} projection and the contribution analysis for \textit{Original labels}, \textit{k-Means} with four clusters ($k$-Means 1), \textit{k-Means} with eight clusters ($k$-Means 2), and \textit{Agglomerative clustering} with nine clusters.

The \textit{Original labels} show that this is a straightforward dataset in terms of class separation. The \textbf{Importance Heatmap} helps us to realize which cluster represents the main area of the dataset: Case-based Reasoning (CBR) in blue (\img{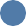}) (see important terms \texttt{reason} and \texttt{case-bas}); Intuitive Logic Programming (ILP) in orange (\img{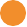}) (see important terms \texttt{program}); and Information Retrieval (IR) in red (\img{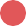}) (see important term \texttt{inform}) -- we highlight these terms in the heatmap. The clustering result with four clusters (\textit{k-Means (1)}) can uncover the information that the smaller cluster is positioned distant from the others since it has no unique influence by specific terms, such as \texttt{case-bas}, \texttt{reason}, \texttt{inform}, or \texttt{program}. The clustering result with eight clusters (\textit{k-Means (2)}) and its respective \textbf{Importance Heatmap} further adds information to the analysis. More terms about Intuitive Logic Program (ILP) appear for the cluster in red (\img{figs/red}) (such as \texttt{logic} and \texttt{program}) -- see these terms in the heatmap highlighted in red. Another important aspect of this clustering result is that the \texttt{learn}, \texttt{inform}, \texttt{retrieve}, \texttt{reason}, and \texttt{case-bas} seem to make cluster \img{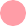} relate with cluster \img{figs/red}. This characteristic is because these terms have similar (with a more significant difference for the term \texttt{learn}) important terms for both clusters.

The third clustering result (\textit{Agglomerative}) shows other interesting characteristics. Here, although only the clusters positioned on the right-bottom of the projections talk about Case-based Reasoning, the \textbf{Importance Heatmap} shows that all clusters contribute from terms of \texttt{case-base} and \texttt{reason}. While these terms contribute to the cluster formation after dimensionality reduction, they could have a negative contribution – meaning that for these clusters not talking about Case-based Reasoning, these terms contribute for these data points to be positioned far away from the bottom-right cluster. The Shapley values representation explain this characteristic for the clusters positioned on top (\img{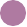}, \img{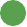}, \img{figs/orange}, and \img{figs/pink}). Each cluster has terms that the presence in the document contributes to its formation. For example, the documents of clusters \img{figs/pink} seem to be influenced by the term \texttt{audio}; the documents of cluster \img{figs/green} seem to talk about information retrieval (see terms \texttt{retrieve} and \texttt{inform}); and so on. However, the dimensionality reduction technique also considered the absence of terms \texttt{case-bas} and \texttt{reason} as an important characteristic for cluster definition.

In this case study, we show that our approach benefits the understanding of dimensionality reduction results after automatically defining clusters. However, as we address in the limitation section (see Section~\ref{sec:limitations}), this is a good approach when the clustering result truly captures the projection's clusters and subclusters.

\section{Discussion}
\label{sec:discussions}

The main contribution of this study consists of explaining dimensionality reduction results using the clusters present in the projected space. Such characteristic is important because the majority of dimensionality reduction techniques are non-linear. While non-linearity is essential for unfolding complex structures in high-dimensional spaces, there is no current way to inverse the calculations and track the features' contributions for generating the projection.

Our methodology provides a correlation between feature values and their importance on cluster results. Such correlation allows analysts to answer questions like ``How increasing/decreasing values for a given feature will influence the classes/clusters in the dataset?'' -- useful information when working with medical datasets, as seen in the case studies. Finally, our coordination mechanism (between the scatter plot and the dot plot) allows identifying intra-cluster patterns.

\subsection{Supporting exploratory analysis}

In exploratory data analysis, analysts want to confirm a hypothesis about a high-dimensional dataset or even want to discover unseen information~\citep{Munzner2015}, making dimensionality reduction techniques an appropriate approach. These aspects make ClusterShapley a valuable tool to interpret dimensionality reduction techniques and help in high-dimensional data analysis. Through the explanations derived using ClusterShapley, analysts can understand various aspects of the data, such as why data samples pertain to a cluster, why different clusters present a relationship, or why data samples of different classes appear very dissimilar in the projected space.

The number of applications for ClusterShapley is numerous, which stresses the contribution of our technique. To cite a few, machine learning practitioners might use ClusterShapley to investigate the quality of feature spaces and understand how different classes appear similar and cause confusion for classification tasks. ClusterShapley can also support annotation of datasets since it gives distinct characteristics about different clusters. Biologists would use ClusterShapley to discover new cell types analyzing the contribution of genes in each cluster visible in the projected space. Finally, ClusterShapley may be useful in applications for monitoring the training process of deep learning models. In this case, an exploratory method using ClusterShapley consists of an early-stopping strategy when the model's learned features investigated through a DR technique and ClusterShapley corresponds to the analyst's mental model.

\subsection{Limitations}
\label{sec:limitations}

\paragraph{\textbf{Run-time execution}} Computing Shapley values is a difficult task. While model-specific strategies~\citep{Lundberg2020} present reasonable execution time, general model implementations such as KernelSHAP~\citep{Lundberg2017} (the one we used in this work) can take too much time to produce explanations for bigger datasets since it uses a weighted linear regression. Thus, we plan to develop approximate strategies for computing Shapley values to explain dimensionality reduction results in future works. For example, one could use a sampling technique that preserves the dataset structures to feed the Shapley estimator. Table~\ref{tab:execution-time} shows how the execution time of estimating Shapley values with KernelSHAP correlates to the size of the dataset, number of clusters, and number of features. Such a characteristic opens the possibility for developing novel strategies that would hierarchically estimate Shapley values. Further that, we plan to develop parallel strategies to estimate Shapley values in the future.

\begin{table}[!h]
\centering
\caption{Time for estimating Shapley values.}
\label{tab:execution-time}
\begin{tabular}{lrrrr}
\toprule 
Dataset & Size & Dim. & Clusters & Time (s)           \\
\midrule 
Iris & 150 & 4 & 3 & 0.1698 \\
Vertebral Column & 310 & 6 & 3 & 0.2555 \\
Indian Liver & 167 & 10 & 6 & 38.22 \\
Red Wine & 1599 & 11 & 10 & 315.71 \\
Com. and Crime & 2215 & 128 & 7 & 462.93 \\
CBR-ILP-IR & 574 & 18694 & 3 & 1098.67 \\
CBR-ILP-IR KM 1 & 574 & 18694 & 4 & 1259.23 \\
CBR-ILP-IR KM 2 & 574 & 18694 & 8 & 1646.52 \\
CBR-ILP-IR Aggl. & 574 & 18694 & 9 & 1659.87 \\
\bottomrule 
\end{tabular}

\end{table}

\paragraph{\textbf{Cluster-oriented analysis}}

We use a cluster-oriented analysis to explain the contribution of features on the data organization in $\mathbb{R}^2$ imposed by dimensionality reduction techniques. That means ClusterShapley returns each feature's contribution to the formation of each cluster of the projected dataset. However, in some cases, it would be useful to understand how features contribute to parts of the projection compared to the remaining of the dataset. While we do not address this scenario, we plan to develop other strategies in future works. For example, through the definition of two clusters (the area of interest and the remaining of the projection), we could compute Shapley values on-the-fly.

Another limitation of our work is related to the quality of the clusters when using by clustering algorithms. In Figure~\ref{fig:clustering-analysis}, the clusters returned by the k-means and agglomerative clustering defined the dataset labels as the first step before using ClusterShapley. Suppose the clusters returned by the clustering algorithm do not convey the real clusters present in the dataset. In that case, ClusterShapley will generate results in which explanations consider dissimilar data points (in different \textit{real} clusters) as similar data points and consequently mislead analysis. Nevertheless, the visualizations provided in this work make it easy for analysts to be aware of these issues.

\section{Conclusion}
\label{sec:conclusion}

Analyzing the clusters imposed by dimensionality reduction techniques is a recurrent task. Understanding which features influenced cluster formation can help discover patterns in data and reveal ubiquitous information. However, most dimensionality reduction techniques are non-linear, which imposes difficulties in tracking the features that contributed to the resulting clusters.

In this work, we use Shapley values to explain the clusters resulted from dimensionality reduction techniques. After defining clusters on visual space, the labels annotate the high-dimensional space to compute each feature's contribution to the projection. From the correlation among contributions and feature values, we discover patterns on medical and social datasets, proving our method's applicability to explain dimensionality reduction results.

In future works, we plan to novel methods to compute Shapley values for dimensionality reduction techniques so that the computation for large sets is not prohibitive. Moreover, we plan to investigate ways to apply dimensionality reduction results in various levels of detail since subclusters provide much insightful information.

\section*{Acknowledgements}

This work was supported by Fundação de Amparo à Pesquisa (FAPESP) [grant numbers \#2018/17881-3, \#2018/25755-8]; the Coordenação de Aperfeiçoamento de Pessoal de Nível Superior (CAPES) [grant number \#88887.487331/2020-00].

\bibliographystyle{cas-model2-names}

\bibliography{cas-refs}

\begin{thebibliography}{46}
\expandafter\ifx\csname natexlab\endcsname\relax\def\natexlab#1{#1}\fi
\providecommand{\url}[1]{\texttt{#1}}
\providecommand{\href}[2]{#2}
\providecommand{\path}[1]{#1}
\providecommand{\DOIprefix}{doi:}
\providecommand{\ArXivprefix}{arXiv:}
\providecommand{\URLprefix}{URL: }
\providecommand{\Pubmedprefix}{pmid:}
\providecommand{\doi}[1]{\href{http://dx.doi.org/#1}{\path{#1}}}
\providecommand{\Pubmed}[1]{\href{pmid:#1}{\path{#1}}}
\providecommand{\bibinfo}[2]{#2}
\ifx\xfnm\relax \def\xfnm[#1]{\unskip,\space#1}\fi
%Type = Article
\bibitem[{Abid et~al.(2018)Abid, Zhang, Bagaria and Zou}]{Abid2018}
\bibinfo{author}{Abid, A.}, \bibinfo{author}{Zhang, M.J.},
  \bibinfo{author}{Bagaria, V.K.}, \bibinfo{author}{Zou, J.},
  \bibinfo{year}{2018}.
\newblock \bibinfo{title}{Exploring patterns enriched in a dataset with
  contrastive principal component analysis}.
\newblock \bibinfo{journal}{Nature communications} \bibinfo{volume}{9},
  \bibinfo{pages}{2134}.
%Type = Book
\bibitem[{Achen(1982)}]{Achen1982}
\bibinfo{author}{Achen, C.}, \bibinfo{year}{1982}.
\newblock \bibinfo{title}{Interpreting and {Using} {Regression}}.
\newblock \bibinfo{address}{Thousand Oaks, California}.
\newblock \DOIprefix\doi{10.4135/9781412984560}.
%Type = Incollection
\bibitem[{Anadón et~al.(2019)Anadón, Martínez-Larrañaga, Ares and
  Martínez}]{Anadon2019}
\bibinfo{author}{Anadón, A.}, \bibinfo{author}{Martínez-Larrañaga, M.R.},
  \bibinfo{author}{Ares, I.}, \bibinfo{author}{Martínez, M.A.},
  \bibinfo{year}{2019}.
\newblock \bibinfo{title}{Chapter 38 - biomarkers of drug toxicity and safety
  evaluation}, in: \bibinfo{editor}{Gupta, R.C.} (Ed.),
  \bibinfo{booktitle}{Biomarkers in Toxicology (Second Edition)}.
  \bibinfo{edition}{second edition} ed.. \bibinfo{publisher}{Academic Press},
  pp. \bibinfo{pages}{655 -- 691}.
%Type = Article
\bibitem[{Bar-Joseph et~al.(2001)Bar-Joseph, Gifford and
  Jaakkola}]{BarJoseph2001}
\bibinfo{author}{Bar-Joseph, Z.}, \bibinfo{author}{Gifford, D.K.},
  \bibinfo{author}{Jaakkola, T.S.}, \bibinfo{year}{2001}.
\newblock \bibinfo{title}{Fast optimal leaf ordering for hierarchical
  clustering}.
\newblock \bibinfo{journal}{Bioinformatics} \bibinfo{volume}{17 Suppl 1},
  \bibinfo{pages}{S22--9}.
%Type = Book
\bibitem[{Bertin(1983)}]{Bertin_1983}
\bibinfo{author}{Bertin, J.}, \bibinfo{year}{1983}.
\newblock \bibinfo{title}{Semiology of Graphics}.
\newblock \bibinfo{publisher}{University of Wisconsin Press}.
%Type = Article
\bibitem[{Coimbra et~al.(2016)Coimbra, Martins, Neves, Telea and
  Paulovich}]{Coimbra2016}
\bibinfo{author}{Coimbra, D.B.}, \bibinfo{author}{Martins, R.M.},
  \bibinfo{author}{Neves, T.T.}, \bibinfo{author}{Telea, A.C.},
  \bibinfo{author}{Paulovich, F.V.}, \bibinfo{year}{2016}.
\newblock \bibinfo{title}{Explaining three-dimensional dimensionality reduction
  plots}.
\newblock \bibinfo{journal}{Information Visualization} \bibinfo{volume}{15},
  \bibinfo{pages}{154--172}.
%Type = Article
\bibitem[{Cortez et~al.(2009)Cortez, Cerdeira, Almeida, Matos and
  Reis}]{Cortez2009}
\bibinfo{author}{Cortez, P.}, \bibinfo{author}{Cerdeira, A.},
  \bibinfo{author}{Almeida, F.}, \bibinfo{author}{Matos, T.},
  \bibinfo{author}{Reis, J.}, \bibinfo{year}{2009}.
\newblock \bibinfo{title}{Modeling wine preferences by data mining from
  physicochemical properties}.
\newblock \bibinfo{journal}{Decision Support Systems} \bibinfo{volume}{47},
  \bibinfo{pages}{547 -- 553}.
\newblock \bibinfo{note}{Smart Business Networks: Concepts and Empirical
  Evidence}.
%Type = Misc
\bibitem[{Davis(2004)}]{Ezekeil2004}
\bibinfo{author}{Davis, E.N.U.}, \bibinfo{year}{2004}.
\newblock \bibinfo{title}{Wine spoilage is legally defined by volatile acidity,
  largely composed of acetic acid}.
\newblock \URLprefix
  \url{https://waterhouse.ucdavis.edu/whats-in-wine/volatile-acidity}.
  \bibinfo{note}{[Online; accessed 01-29-2020]}.
%Type = Incollection
\bibitem[{Dhillon and Steadman(2012)}]{Dhillon2012}
\bibinfo{author}{Dhillon, A.}, \bibinfo{author}{Steadman, R.H.},
  \bibinfo{year}{2012}.
\newblock \bibinfo{title}{Chapter 5 - liver diseases}, in:
  \bibinfo{editor}{Fleisher, L.A.} (Ed.), \bibinfo{booktitle}{Anesthesia and
  Uncommon Diseases (Sixth Edition)}. \bibinfo{edition}{sixth edition} ed..
  \bibinfo{publisher}{W.B. Saunders}, \bibinfo{address}{Philadelphia}, pp.
  \bibinfo{pages}{162 -- 214}.
%Type = Misc
\bibitem[{Dua and Graff(2017)}]{Dua2019}
\bibinfo{author}{Dua, D.}, \bibinfo{author}{Graff, C.}, \bibinfo{year}{2017}.
\newblock \bibinfo{title}{{UCI} machine learning repository}.
\newblock \URLprefix \url{http://archive.ics.uci.edu/ml}.
%Type = Article
\bibitem[{Eler et~al.(2018)Eler, Grosa, Pola, Garcia, Correia and
  Teixeira}]{Eler2018}
\bibinfo{author}{Eler, D.M.}, \bibinfo{author}{Grosa, D.},
  \bibinfo{author}{Pola, I.}, \bibinfo{author}{Garcia, R.},
  \bibinfo{author}{Correia, R.}, \bibinfo{author}{Teixeira, J.},
  \bibinfo{year}{2018}.
\newblock \bibinfo{title}{Analysis of document pre-processing effects in text
  and opinion mining}.
\newblock \bibinfo{journal}{Information} \bibinfo{volume}{9}.
\newblock \URLprefix \url{https://www.mdpi.com/2078-2489/9/4/100},
  \DOIprefix\doi{10.3390/info9040100}.
%Type = Article
\bibitem[{Fujiwara et~al.(2019)Fujiwara, Kwon and Ma}]{Fujiwara2019}
\bibinfo{author}{Fujiwara, T.}, \bibinfo{author}{Kwon, O.H.},
  \bibinfo{author}{Ma, K.L.}, \bibinfo{year}{2019}.
\newblock \bibinfo{title}{Supporting analysis of dimensionality reduction
  results with contrastive learning}.
\newblock \bibinfo{journal}{IEEE Trans. Vis. and Comp. Graph.}
  \bibinfo{volume}{26}, \bibinfo{pages}{45--55}.
%Type = Article
\bibitem[{Goyal and May(2017)}]{ALT}
\bibinfo{author}{Goyal, H.}, \bibinfo{author}{May, E.}, \bibinfo{year}{2017}.
\newblock \bibinfo{title}{Roadmap for evaluation of abnormal liver
  chemistries}.
\newblock \bibinfo{journal}{Journal of Laboratory and Precision Medicine}
  \bibinfo{volume}{2}.
%Type = Inbook
\bibitem[{Izenman(2008)}]{Izenman2008}
\bibinfo{author}{Izenman, A.J.}, \bibinfo{year}{2008}.
\newblock \bibinfo{title}{Linear Discriminant Analysis}.
  \bibinfo{publisher}{Springer New York}, \bibinfo{address}{New York, NY}.
\newblock pp. \bibinfo{pages}{237--280}.
%Type = Article
\bibitem[{Joia et~al.(2015)Joia, Petronetto and Nonato}]{Joia2015}
\bibinfo{author}{Joia, P.}, \bibinfo{author}{Petronetto, F.},
  \bibinfo{author}{Nonato, L.}, \bibinfo{year}{2015}.
\newblock \bibinfo{title}{Uncovering representative groups in multidimensional
  projections}.
\newblock \bibinfo{journal}{CGF} \bibinfo{volume}{34},
  \bibinfo{pages}{281--290}.
%Type = Book
\bibitem[{Kaufman and Rousseuw(2005)}]{KaufmanRousseuw2005}
\bibinfo{author}{Kaufman, L.}, \bibinfo{author}{Rousseuw, P.J.},
  \bibinfo{year}{2005}.
\newblock \bibinfo{title}{Finding Groups in Data: An Introduction to Cluster
  Analysis}.
\newblock \bibinfo{publisher}{Wiley-Interscience}, \bibinfo{address}{Principles
  and Practice}.
%Type = Book
\bibitem[{Kruskal and Wish(1978)}]{Kruskal1978}
\bibinfo{author}{Kruskal, J.}, \bibinfo{author}{Wish, M.},
  \bibinfo{year}{1978}.
\newblock \bibinfo{title}{{Multidimensional Scaling}}.
\newblock \bibinfo{publisher}{Sage Publications}.
%Type = Article
\bibitem[{Kwon et~al.(2018)Kwon, Eysenbach, Verma, Ng, Filippi, Stewart and
  Perer}]{Kwon2018}
\bibinfo{author}{Kwon, B.}, \bibinfo{author}{Eysenbach, B.},
  \bibinfo{author}{Verma, J.}, \bibinfo{author}{Ng, K.},
  \bibinfo{author}{Filippi, C.D.}, \bibinfo{author}{Stewart, W.F.},
  \bibinfo{author}{Perer, A.}, \bibinfo{year}{2018}.
\newblock \bibinfo{title}{Clustervision: Visual supervision of unsupervised
  clustering}.
\newblock \bibinfo{journal}{IEEE Trans. Vis. Comput. Graph.}
  \bibinfo{volume}{24}, \bibinfo{pages}{142--151}.
%Type = Article
\bibitem[{Labelle et~al.(2005)Labelle, Roussouly, Berthonnaud, Dimnet and
  O'Brien}]{Labelle2005}
\bibinfo{author}{Labelle, H.}, \bibinfo{author}{Roussouly, P.},
  \bibinfo{author}{Berthonnaud, E.}, \bibinfo{author}{Dimnet, J.},
  \bibinfo{author}{O'Brien, M.}, \bibinfo{year}{2005}.
\newblock \bibinfo{title}{The importance of spino-pelvic balance in l5-s1
  developmental spondylolisthesis: A review of pertinent radiologic
  measurements}.
\newblock \bibinfo{journal}{Spine} \bibinfo{volume}{30},
  \bibinfo{pages}{27--34}.
%Type = Misc
\bibitem[{Lowe and John(2018)}]{Lowe2018}
\bibinfo{author}{Lowe, D.}, \bibinfo{author}{John, S.}, \bibinfo{year}{2018}.
\newblock \bibinfo{title}{Alkaline phosphatase}.
\newblock \URLprefix \url{https://www.ncbi.nlm.nih.gov/books/NBK459201/}.
  \bibinfo{note}{[Online; accessed 01-29-2020]}.
%Type = Article
\bibitem[{Lundberg et~al.(2020)Lundberg, Erion, Chen, DeGrave, Prutkin, Nair,
  Katz, Himmelfarb, Bansal and Lee}]{Lundberg2020}
\bibinfo{author}{Lundberg, S.M.}, \bibinfo{author}{Erion, G.},
  \bibinfo{author}{Chen, H.}, \bibinfo{author}{DeGrave, A.},
  \bibinfo{author}{Prutkin, J.M.}, \bibinfo{author}{Nair, B.},
  \bibinfo{author}{Katz, R.}, \bibinfo{author}{Himmelfarb, J.},
  \bibinfo{author}{Bansal, N.}, \bibinfo{author}{Lee, S.I.},
  \bibinfo{year}{2020}.
\newblock \bibinfo{title}{From local explanations to global understanding with
  explainable ai for trees}.
\newblock \bibinfo{journal}{Nature Machine Intelligence} \bibinfo{volume}{2},
  \bibinfo{pages}{2522--5839}.
%Type = Incollection
\bibitem[{Lundberg and Lee(2017)}]{Lundberg2017}
\bibinfo{author}{Lundberg, S.M.}, \bibinfo{author}{Lee, S.I.},
  \bibinfo{year}{2017}.
\newblock \bibinfo{title}{A unified approach to interpreting model
  predictions}, in: \bibinfo{editor}{Guyon, I.}, \bibinfo{editor}{Luxburg,
  U.V.}, \bibinfo{editor}{Bengio, S.}, \bibinfo{editor}{Wallach, H.},
  \bibinfo{editor}{Fergus, R.}, \bibinfo{editor}{Vishwanathan, S.},
  \bibinfo{editor}{Garnett, R.} (Eds.), \bibinfo{booktitle}{Advances in Neural
  Information Processing Systems 30}, pp. \bibinfo{pages}{4765--4774}.
%Type = Article
\bibitem[{Lähnemann et~al.(2020)Lähnemann, Köster and
  Szczurek}]{Lahnemann2020}
\bibinfo{author}{Lähnemann, D.}, \bibinfo{author}{Köster, J.},
  \bibinfo{author}{Szczurek, E.e.a.}, \bibinfo{year}{2020}.
\newblock \bibinfo{title}{Eleven grand challenges in single-cell data science}.
\newblock \bibinfo{journal}{Genome Biol} \bibinfo{volume}{31}.
%Type = Article
\bibitem[{Maaten and Hinton(2008)}]{Maaten_2008}
\bibinfo{author}{Maaten, L.J.P.}, \bibinfo{author}{Hinton, G.E.},
  \bibinfo{year}{2008}.
\newblock \bibinfo{title}{Visualizing high-dimensional data using t-sne}.
\newblock \bibinfo{journal}{Journal of Machine Learning Research}
  \bibinfo{volume}{9}, \bibinfo{pages}{2579--–2605}.
%Type = Article
\bibitem[{Marcilio et~al.(2017)Marcilio, Eler and Garcia}]{MarcilioJr2017}
\bibinfo{author}{Marcilio, W.E.}, \bibinfo{author}{Eler, D.M.},
  \bibinfo{author}{Garcia, R.E.}, \bibinfo{year}{2017}.
\newblock \bibinfo{title}{An approach to perform local analysis on
  multidimensional projection}.
\newblock \bibinfo{journal}{30th SIBGRAPI Conf. on Graph., Patterns and Images
  (SIBGRAPI)} , \bibinfo{pages}{351--358}.
%Type = Article
\bibitem[{Marcilio-Jr et~al.(2020)Marcilio-Jr, Eler, Garcia, Correia and
  Silva}]{Marcilio2020}
\bibinfo{author}{Marcilio-Jr, W.}, \bibinfo{author}{Eler, D.},
  \bibinfo{author}{Garcia, R.}, \bibinfo{author}{Correia, R.},
  \bibinfo{author}{Silva, L.F.}, \bibinfo{year}{2020}.
\newblock \bibinfo{title}{A hybrid visualization approach to perform analysis
  of feature spaces}.
\newblock \bibinfo{journal}{International Conference on Information
  Technology–New Generations} \bibinfo{volume}{1134}.
%Type = Article
\bibitem[{{McInnes} et~al.(2018){McInnes}, {Healy} and
  {Melville}}]{McInnes2018}
\bibinfo{author}{{McInnes}, L.}, \bibinfo{author}{{Healy}, J.},
  \bibinfo{author}{{Melville}, J.}, \bibinfo{year}{2018}.
\newblock \bibinfo{title}{{UMAP: Uniform Manifold Approximation and Projection
  for Dimension Reduction}}.
\newblock \bibinfo{journal}{ArXiv e-prints}
  \href{http://arxiv.org/abs/1802.03426}{\tt arXiv:1802.03426}.
%Type = Book
\bibitem[{Molnar(2019)}]{Molnar2019}
\bibinfo{author}{Molnar, C.}, \bibinfo{year}{2019}.
\newblock \bibinfo{title}{Interpretable Machine Learning}.
\newblock
  \bibinfo{note}{\url{https://christophm.github.io/interpretable-ml-book/}}.
%Type = Book
\bibitem[{Munzner(2015)}]{Munzner2015}
\bibinfo{author}{Munzner, T.}, \bibinfo{year}{2015}.
\newblock \bibinfo{title}{Visualization Analysis and Design}.
\newblock AK Peters Visualization Series, \bibinfo{publisher}{CRC Press}.
\newblock \URLprefix \url{https://books.google.de/books?id=NfkYCwAAQBAJ}.
%Type = Article
\bibitem[{Müllner(2011)}]{Mullner2011}
\bibinfo{author}{Müllner, D.}, \bibinfo{year}{2011}.
\newblock \bibinfo{title}{Modern hierarchical, agglomerative clustering
  algorithms}.
\newblock \bibinfo{journal}{CoRR} \bibinfo{volume}{abs/1109.2378}.
%Type = Inproceedings
\bibitem[{Pagliosa et~al.(2016)Pagliosa, Pagliosa and Nonato}]{Pagliosa2016}
\bibinfo{author}{Pagliosa, L.C.}, \bibinfo{author}{Pagliosa, P.A.},
  \bibinfo{author}{Nonato, L.G.}, \bibinfo{year}{2016}.
\newblock \bibinfo{title}{Understanding attribute variability in
  multidimensional projections}, in: \bibinfo{booktitle}{29th Conf. Graphics,
  Patterns and Images (SIBGRAPI)}, pp. \bibinfo{pages}{297--304}.
%Type = Article
\bibitem[{Parzen(1962)}]{Parzen1962}
\bibinfo{author}{Parzen, E.}, \bibinfo{year}{1962}.
\newblock \bibinfo{title}{On estimation of a probability density function and
  mode}.
\newblock \bibinfo{journal}{The Annals of Mathematical Statistics}
  \bibinfo{volume}{33}, \bibinfo{pages}{1065--1076}.
%Type = Article
\bibitem[{Paulovich et~al.(2008)Paulovich, Nonato, Rosane and
  Levkowitz}]{Paulovich_2008}
\bibinfo{author}{Paulovich, F.V.}, \bibinfo{author}{Nonato, L.G.},
  \bibinfo{author}{Rosane, M.}, \bibinfo{author}{Levkowitz, H.},
  \bibinfo{year}{2008}.
\newblock \bibinfo{title}{Least square projection: A fast high-precision
  multidimensional projection technique and its application to document
  mapping}.
\newblock \bibinfo{journal}{IEEE Transactions on Visulization and Computer
  Graphics} \bibinfo{volume}{3}, \bibinfo{pages}{564--575}.
%Type = Article
\bibitem[{Pezzotti et~al.(2018)Pezzotti, H{\"o}llt, van Gemert, Lelieveldt,
  Eisemann and Vilanova}]{Pezzotti2018}
\bibinfo{author}{Pezzotti, N.}, \bibinfo{author}{H{\"o}llt, T.},
  \bibinfo{author}{van Gemert, J.}, \bibinfo{author}{Lelieveldt, B.},
  \bibinfo{author}{Eisemann, E.}, \bibinfo{author}{Vilanova, A.},
  \bibinfo{year}{2018}.
\newblock \bibinfo{title}{Deepeyes: Progressive visual analytics for designing
  deep neural networks}.
\newblock \bibinfo{journal}{IEEE Transactions on Visualization and Computer
  Graphics (Proceedings of IEEE VAST 2017)} \bibinfo{volume}{24},
  \bibinfo{pages}{98 -- 108}.
\newblock \DOIprefix\doi{10.1109/TVCG.2017.2744358}.
%Type = Article
\bibitem[{Redmond and Baveja(2002)}]{Redmon2002}
\bibinfo{author}{Redmond, M.}, \bibinfo{author}{Baveja, A.},
  \bibinfo{year}{2002}.
\newblock \bibinfo{title}{A data-driven software tool for enabling cooperative
  information sharing among police departments.}
\newblock \bibinfo{journal}{European Journal of Operational Research}
  \bibinfo{volume}{141}, \bibinfo{pages}{660--678}.
%Type = Article
\bibitem[{Rosenblatt(1956)}]{Rosenblatt1956}
\bibinfo{author}{Rosenblatt, M.}, \bibinfo{year}{1956}.
\newblock \bibinfo{title}{Remarks on some nonparametric estimates of a density
  function}.
\newblock \bibinfo{journal}{Annals of Mathematical Statistics}
  \bibinfo{volume}{27}, \bibinfo{pages}{832--837}.
%Type = Article
\bibitem[{Roussouly and Pinheiro-Fraco(2011)}]{Roussouly2011}
\bibinfo{author}{Roussouly, P.}, \bibinfo{author}{Pinheiro-Fraco, J.},
  \bibinfo{year}{2011}.
\newblock \bibinfo{title}{Biomechanical analysis of the spino-pelvic
  organization and adaptation in pathology}.
\newblock \bibinfo{journal}{Eur Spine Journal} .
%Type = Article
\bibitem[{Shapley(1953)}]{Shapley1953}
\bibinfo{author}{Shapley, L.}, \bibinfo{year}{1953}.
\newblock \bibinfo{title}{A value for n-person games, vol ii of contributions
  to the theory of games} .
%Type = Inproceedings
\bibitem[{Silva et~al.(2015)Silva, Rauber, Martins, Minghim and
  Telea}]{Silva2015}
\bibinfo{author}{Silva, R.R.O.d.}, \bibinfo{author}{Rauber, P.E.},
  \bibinfo{author}{Martins, R.M.}, \bibinfo{author}{Minghim, R.},
  \bibinfo{author}{Telea, A.C.}, \bibinfo{year}{2015}.
\newblock \bibinfo{title}{{Attribute-based Visual Explanation of
  Multidimensional Projections}}, in: \bibinfo{editor}{Bertini, E.},
  \bibinfo{editor}{Roberts, J.C.} (Eds.), \bibinfo{booktitle}{EuroVis Workshop
  on Visual Analytics (EuroVA)}.
%Type = Article
\bibitem[{Stahnke et~al.(2016)Stahnke, D{\"o}rk, M{\"u}ller and
  Thom}]{Stahnke2016}
\bibinfo{author}{Stahnke, J.}, \bibinfo{author}{D{\"o}rk, M.},
  \bibinfo{author}{M{\"u}ller, B.}, \bibinfo{author}{Thom, A.},
  \bibinfo{year}{2016}.
\newblock \bibinfo{title}{Probing projections: Interaction techniques for
  interpreting arrangements and errors of dimensionality reductions}.
\newblock \bibinfo{journal}{IEEE Trans. on Vis. and Comp. Graph.}
  \bibinfo{volume}{22}, \bibinfo{pages}{629--638}.
%Type = Article
\bibitem[{Targher and Byrne(2015)}]{ALP}
\bibinfo{author}{Targher, G.}, \bibinfo{author}{Byrne, C.},
  \bibinfo{year}{2015}.
\newblock \bibinfo{title}{Circulating markers of liver function and
  cardiovascular disease risk}.
\newblock \bibinfo{journal}{Arteriosclerosis, Thrombosis, and Vascular Biology}
  \bibinfo{volume}{35}, \bibinfo{pages}{2290–2296}.
%Type = Article
\bibitem[{Tebet(2014)}]{Tebet2013}
\bibinfo{author}{Tebet, M.}, \bibinfo{year}{2014}.
\newblock \bibinfo{title}{Current concepts on the sagittal balance and
  classification of spondylolysis and spondylolisthesis}.
\newblock \bibinfo{journal}{Rev Bras Ortop} , \bibinfo{pages}{3--12}.
%Type = Article
\bibitem[{Turkay et~al.(2012)Turkay, Lundervold, Lundervold and
  Hauser}]{Turkay2012}
\bibinfo{author}{Turkay, C.}, \bibinfo{author}{Lundervold, A.},
  \bibinfo{author}{Lundervold, A.J.}, \bibinfo{author}{Hauser, H.},
  \bibinfo{year}{2012}.
\newblock \bibinfo{title}{Representative factor generation for the interactive
  visual analysis of high-dimensional data}.
\newblock \bibinfo{journal}{{IEEE} Trans. Vis. Comput. Graph.}
  \bibinfo{volume}{18}, \bibinfo{pages}{2621--2630}.
%Type = Article
\bibitem[{van Unen et~al.(2018)van Unen, Höllt and Pezzotti}]{Unen2018}
\bibinfo{author}{van Unen, V.}, \bibinfo{author}{Höllt, T.},
  \bibinfo{author}{Pezzotti, N.e.a.}, \bibinfo{year}{2018}.
\newblock \bibinfo{title}{Visual analysis of mass cytometry data by
  hierarchical stochastic neighbour embedding reveals rare cell types}.
\newblock \bibinfo{journal}{Nat Commun}
  \DOIprefix\doi{10.1038/s41467-017-01689-9}.
%Type = Article
\bibitem[{\v{S}trumbelj and Kononenko(2014)}]{Strumbelj2014}
\bibinfo{author}{\v{S}trumbelj, E.}, \bibinfo{author}{Kononenko, I.},
  \bibinfo{year}{2014}.
\newblock \bibinfo{title}{Explaining prediction models and individual
  predictions with feature contributions}.
\newblock \bibinfo{journal}{Knowl. Inf. Syst.} \bibinfo{volume}{41},
  \bibinfo{pages}{647–665}.
%Type = Article
\bibitem[{Wang et~al.(2017)Wang, Li, Nie, Theisel, Gong and Lehmann}]{Wang2017}
\bibinfo{author}{Wang, Y.}, \bibinfo{author}{Li, J.}, \bibinfo{author}{Nie,
  F.}, \bibinfo{author}{Theisel, H.}, \bibinfo{author}{Gong, M.},
  \bibinfo{author}{Lehmann, D.J.}, \bibinfo{year}{2017}.
\newblock \bibinfo{title}{Linear discriminative star coordinates for exploring
  class and cluster separation of high dimensional data}.
\newblock \bibinfo{journal}{Computer Graphics Forum} \bibinfo{volume}{36},
  \bibinfo{pages}{401--410}.

\end{thebibliography}

% \bio{figs/pic1} BLIND

\end{document}